\documentclass[twocolumn]{IEEEtran}
\usepackage{lscape}
\usepackage{booktabs}
\usepackage{authblk}
\usepackage{url}
\usepackage{silence}
\usepackage[font=scriptsize]{caption}

\usepackage[pdftex]{graphicx}
\usepackage{epstopdf}
\usepackage{pifont}
\usepackage[table,xcdraw]{xcolor}
\usepackage{array}
\usepackage{verse}
\usepackage{longtable}
\usepackage{supertabular}
\usepackage{cite}
\usepackage{color}
\usepackage{amsmath}
\usepackage{lettrine}
\usepackage{hyperref}
\usepackage{multirow}
\usepackage{multicol}
\usepackage{dirtytalk}
\usepackage{color}

\hypersetup{
    colorlinks=false,
    linkcolor=blue,
    filecolor=magenta,
    urlcolor=cyan,
}

\begin{document}

\title{Security of Future Self-Driving Vehicular Networking Systems: Challenges and Solutions}
\title{Securing Self-Driving Cars and ML-Driven VANETs: The Challenges and The Way Forward}
\title{Securing Future Autonomous \& Connected Vehicles: Challenges Posed by Adversarial Machine Learning and The Way Forward}
\title{Securing Connected \& Autonomous Vehicles: Challenges Posed by Adversarial Machine Learning and The Way Forward}
%\title{Securing Future Autonomous/Connected Vehicles: Challenges Posed by Adversarial Machine Learning and The Way Forward}

\author{Adnan Qayyum$^1$}
\author{Muhammad Usama$^1$}
\author{Junaid Qadir$^1$}
\author{Ala Al-Fuqaha$^2$}
\affil{$^1$ Information Technology University (ITU), Punjab, Lahore, Pakistan \newline
$^2$ Hamad Bin Khalifa University (HBKU), Doha, Qatar}
\maketitle

%\jq{Write the abstract. Make sure that you introduce the area well and highlight the contributions of this paper and how it extends the state of the art}
%\jq{I have changed the title to reflect the specific focus of our magazine article. The bulk of the paper should focus on the new challenges induced by Adversarial ML. We will provide sufficient background and highlight existing challenges but the main focus should be on adversarial ML. Please reflect this by rewriting the abstract, conclusions, and the rest of the paper.}
% Vehicular ad hoc networks (VANETs) \jq{use the standard terms CAVs as well in the abstract}

\begin{abstract}
Connected and autonomous vehicles (CAVs) will form the backbone of future next-generation intelligent transportation systems (ITS) providing travel comfort, road safety, along with a number of value-added services. Such a transformation---which will be fuelled by concomitant advances in technologies for machine learning (ML) and wireless communications---will enable a future vehicular ecosystem that is better featured and more efficient. However, there are lurking security problems related to the use of ML in such a critical setting where an incorrect ML decision may not only be a nuisance but can lead to loss of precious lives. In this paper, we present an in-depth overview of the various challenges associated with the application of ML in vehicular networks. In addition, we formulate the ML pipeline of CAVs and present various potential security issues associated with the adoption of ML methods. In particular, we focus on the perspective of adversarial ML attacks on CAVs and outline a solution to defend against adversarial attacks in multiple settings.
\end{abstract}

\begin{IEEEkeywords}
Connected and autonomous vehicles, machine/deep learning, adversarial machine learning, adversarial perturbation, perturbation
detection, and robust machine learning.
\end{IEEEkeywords}

\section{Introduction}
\label{sec:intro}
%--- Why is the subject area important?

%The vehicular ad hoc networks (VANETs) are becoming increasingly important with the development of vehicle industry, wireless communications technologies, and artificial intelligence (AI) research. 
% \textcolor{red}{ALA: I suggest that we keep the introduction concise to capture the interest of the readers and provide overall outline of the paper and how it is different from the state-of-the-art surveys. All other material can go to the background section.} \textcolor{red}{ALA: As a comment on the whole survey, it is critical to provide references when needed and not to leave any material that is PURELY adversarial machine learning because the literature if full of it. Therefore, it is important to tie everything back to connected and autonomous vehicles as this is the contribution of this survey. This is a vertically-focused survey on adversarial machine learning in the context of CAVs.}

In recent years, connected and autonomous vehicles (CAVs) have emerged as a promising area of research. The connected vehicles are an important component of intelligent transportation systems (ITS) in which vehicles communicate with each other and with communications infrastructure to exchange safety messages and other critical information (e.g., traffic and road conditions). One of the main driving force for CAVs is the advancement of machine learning (ML) methods, particularly deep learning (DL), that are used for decision making at different levels. Unlike conventional connected vehicles, the autonomous (self-driving) vehicles have two important characteristics; namely, automation capability and cooperation (connectivity) \cite{shladover2018connected}. In future smart cities, CAVs are expected to have a profound impact on the vehicular ecosystem and society.   

%\jq{Talk about the role of ML in building up autonomous vehicles and connected vehicles. The difference between autonomous vehicles and connected vehicles should be clarified: cite Connected and automated vehicle systems: Introduction and overview \cite{shladover2018connected}.}
The phenomenon of connected vehicles is realized through technology known as vehicular networks or vehicular ad-hoc networks (VANETs) \cite{hussain2018autonomous}. Over the years, various configurations of connected vehicles have been developed including the use of dedicated short-range communications (DSRC) in the United States and ITS-G5 in Europe based on the IEEE 802.11p standard. However, a recent study  \cite{araniti2013lte} has shown many limitations of such systems such as (1) short-lived infrastructure-to-vehicle (I2V) connection, (2) non-guaranteed quality of service (QoS), and (3) unbounded channel access delay, etc. To address such limitations, the 3rd generation partnership project (3GPP) has been initiated with a focus on leveraging the high penetration rate of long term evolution (LTE) and 5G cellular networks to support vehicle-to-everything (V2X) services \cite{peng2018vehicular}. The purpose of developing V2X technology is to enable the communication between all entities encountered in the road environment including vehicles, communications infrastructure, pedestrians, cycles, etc. 

%\jq{In this paragraph, highlight the important role ML will play in CAVs.} 
The impressive ability of ML/DL to leverage increasingly accessible data, along with the advancement in other concomitant technologies (such as wireless communications), seems to be set to enable autonomous and self-organizing connected vehicles in the future. In addition, future vehicular networks will evolve from normal to autonomous vehicles and will enable ubiquitous Internet access on vehicles. ML will have a predominant role in building the perception system of autonomous and semi-autonomous connected vehicles.

\begin{figure*}
\begin{center}
\scriptsize
\includegraphics[width=.9\textwidth]{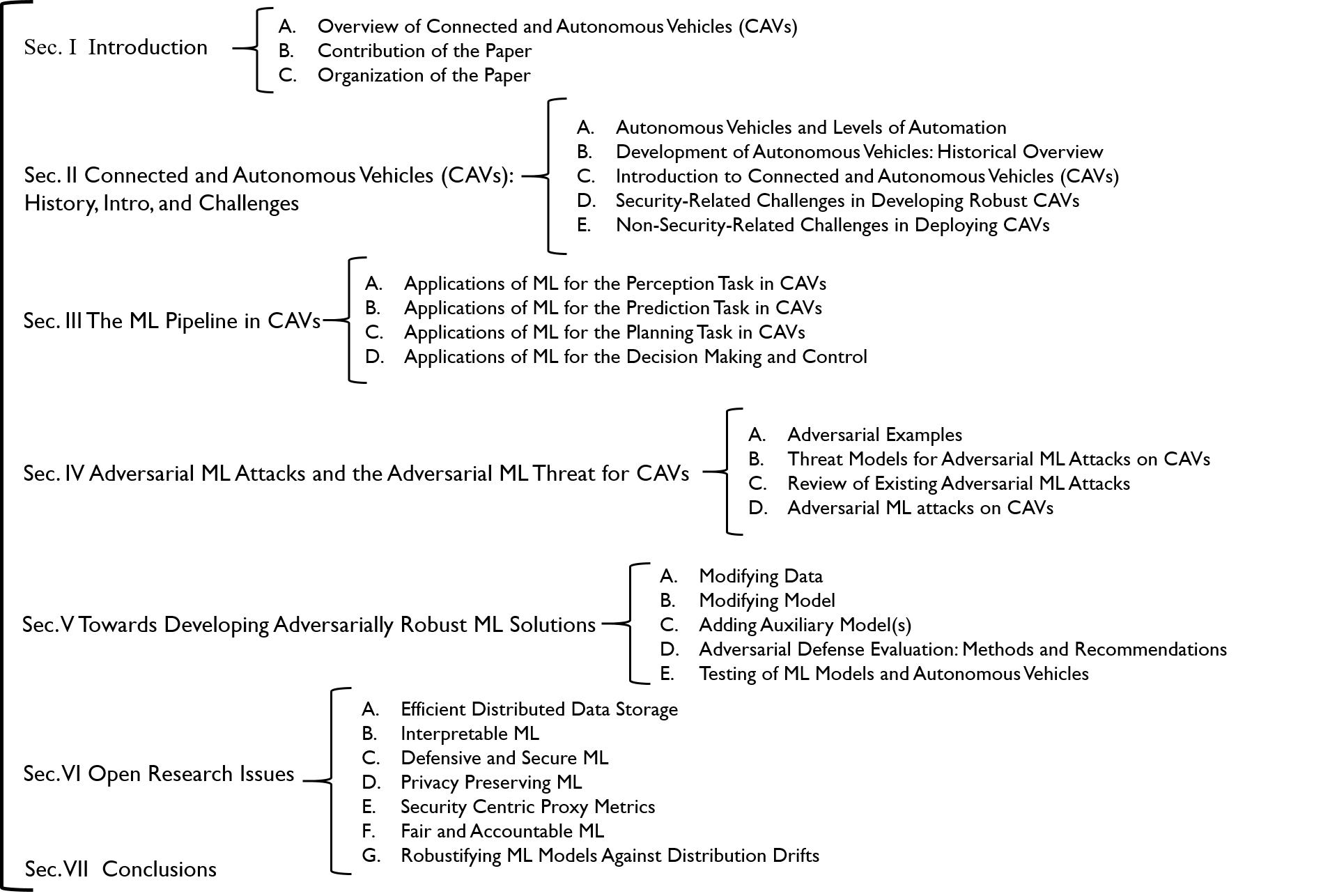}
\caption{Outline of the paper}
\label{fig:outline}
\end{center}
\end{figure*}

%(Cite some more details and refer to a reference that includes more details—You can learn more by seeing the works of Shladover, Steven E).

% The current formulation of CAVs is essentially using ML/DL methods at a large scale

%To address such challenges various approaches have been presented in literature including ML/DL-based methods.

% \vspace{2mm}
% \textit{Different Configurations of VANETs}

% Toward cloud-based vehicular networks with efficient resource management \cite{yu2013toward}
% Broadcasting safety information in vehicular networks: issues and approaches \cite{chen2010broadcasting};
% Heterogeneous vehicular networking: A survey on architecture, challenges, and solutions \cite{zheng2015heterogeneous}

%The security of VANETs and autonomous cars has emerged as an open research problem in recent years. 
%Although different VANETs systems have been developed over the time, they are still vulnerable to various security issues. 

Despite the development of different configurations of connected vehicles, they are still vulnerable to various security issues and there are various automotive attack surfaces that can be exploited \cite{checkoway2011comprehensive}. The threat is getting worse with the development of fully autonomous vehicles. As the autonomous vehicles are being equipped with many sensors such as cameras, RADAR, LIDAR, and mechanical control units, etc. These sensors share critical sensory information with onboard devices through CAN bus and with other nearby vehicles as well. The backbone of self-driving vehicles is the onboard intelligent processing capabilities using the data collected through the sensory system. This data can be used for many other purposes, e.g., getting information about vehicle kinetics, traffic flow, road, and network conditions, etc. Such data can be potentially used for improving the performance of the vehicular ecosystem using adaptive data-driven decision making and can also be used to accomplish various destructive objectives. Therefore, ensuring data integrity and security are necessarily important to avoid various risks and attacks on CAVs.

It is common for the perception and control systems of CAVs to be built using ML/DL methods. However, ML/DL techniques have been recently found vulnerable to carefully crafted adversarial perturbations \cite{papernot2016limitations} and different physical world attacks have been successfully performed on the vision system of autonomous cars \cite{roadsigns17,sitawarin2018darts}. This has raised many privacy and security concerns about the use of such methods particularly for security-critical applications like CAVs. In this paper, we aim to highlight various security issues associated with the use of ML and we present a review of adversarial ML literature mainly focusing on CAVs. In addition, we also present a taxonomy of possible solutions to restrict adversarial ML attacks and open research issues on autonomous, connected vehicles, and ML.  

ML in general and DL schemes specifically perform exceptionally well in learning hidden patterns from data. DL schemes such as deep neural networks (DNN) have outperformed human-level intelligence in many perception and detection tasks by accurately learning from a large corpus of training data and classifying/predicting with high accuracy on unseen real-world test examples. As DL schemes produce outstanding results, they have been used in many real-world security-sensitive tasks such as perception system in self-driving cars, anomaly and intrusion detection in vehicular networks, etc. ML/DL schemes are designed for benign and stationary environments where it is assumed that the training and test data belongs to the same statistical distribution. The application of this assumption in a real-world application is flawed as training and test data can have different statistical distributions which gives rise to an opening for adversaries to compromise the ML/DL-based systems. Furthermore, the lack of interpretability of the learning process, imperfections in training process, and discontinuity in the input-output relationship of DL schemes also resulted in an incentive for adversaries to fool the deployed ML/DL system \cite{szegedy2013intriguing}.

\begin{table*}[]
\centering
\scriptsize
\caption{Comparison of this paper with existing survey and review papers on the security of machine learning (ML) and connected and autonomous vehicles (CAVs). (Legend:$\surd$ means covered;$\times$ means not covered; $\approx$ means partially covered.)}
\scalebox{.9}{
\begin{tabular}{|p{5mm}|p{13mm}|p{22mm}|p{7mm}|p{22mm}|p{12mm}|p{10mm}|p{12mm}|p{10mm}|p{12mm}|p{10mm}|p{10mm}|}
\hline
%\#Refs
\textbf{Year} & \textbf{Authors} &  \textbf{Publisher} & \textbf{Papers Cited} & \textbf{Focused Area} & \textbf{Conventional Challenges} & \textbf{Threat Models} & \textbf{Adversarial ML} & \textbf{Robust ML Solutions} & \textbf{Autonomous Vehicles} & \textbf{Connected Vehicles} & \textbf{Open Research Issues} \\ \hline
2014 &  Mejri et al. \cite{mejri2014survey} & Elsevier Vehicular Communication & 69 & Security of vehicular networks  & $\surd$ & $\times$ & $\times$ & $\times$ & $\times$ & $\surd$ & $\times$ \\ \hline 
2016 & Gardiner et al. \cite{gardiner2016security} & ACM Computing Surveys (CSUR) & 40 & Security of ML for malware classification & $\times$  & $\surd$ & $\surd$  & $\approx$ & $\times$ & $\times$ & $\times$ \\ \hline
2018 & Chakraborty et al. \cite{chakraborty2018adversarial} & arXiv & 79 & Adversarial attacks and defenses & $\times$ & $\surd$ & $\surd$ & $\surd$ & $\times$ & $\times$ & $\times$ \\ \hline
2018 & Akhter et al. \cite{akhtar2018threat} & IEEE Access & 195 & Adversarial attacks and defenses in computer vision & $\times$ & $\times$  & $\surd$ & $\surd$ & $\approx$ & $\times$ & $\times$ \\ \hline
2018 & Siegel et al. \cite{siegel2018survey} & IEEE Transactions on ITS & 198 & Survey on connected vehicles' landscape & $\surd$  & $\times$  & $\times$  & $\times$ & $\times$ & $\surd$ & $\times$ \\ \hline
2018 & Hussain et al. \cite{hussain2018autonomous} & IEEE Communications Surveys and Tutorials (COMST) & 230 & Autonomous cars: research results, issues and future challenges & $\surd$  & $\times$  & $\times$ & $\times$ & $\surd$ & $\times$ & $\surd$\\ \hline
2019 & Yuan et al. \cite{yuan2019adversarial} & IEEE Transactions on NN \& LS (TNNLS) &  146 & Adversarial attacks and defenses for deep learning systems & $\times$ & $\surd$ & $\surd$ & $\surd$ & $\times$ & $\times$ & $\times$ \\ \hline
2019 & Wang et al. \cite{wang2019survey} & arXiv &  128  & Adversarial ML attacks and defenses in text domain & $\times$ & $\surd$  & $\surd$ & $\surd$ & $\times$ & $\times$ & $\times$ \\ \hline
2019 & \multicolumn{2}{c|}{Our Paper} & 239 & Security of CAVs and ML  & $\surd$ & $\surd$ & $\surd$ & $\surd$ & $\surd$ & $\surd$ & $\surd$  \\ \hline
\end{tabular}}
\end{table*}

%\vspace{2mm}
\textit{Contributions of this Paper}: %\textit{Why is this survey timely?} \jq{Important to write this well!}
In this paper, we build upon the existing literature available on CAVs and present a comprehensive review of that literature. The following are the major contributions of this study. 

\begin{enumerate}
    %\item We present a timeline for the development of autonomous vehicles;
    \item We formulate the ML pipeline of CAVs and describe in detail various security challenges that arise with the increasing adoption of ML techniques in CAVs, specifically emphasizing the challenges posed by adversarial ML;  
    \item We present a taxonomy of various threat models and highlight the generalization of attack surfaces for general ML, autonomous, and connected vehicle applications;
    \item We review existing adversarial ML attacks with a special emphasis on their relevance for CAVs;
    \item We review robust ML approaches and provide a taxonomy of these approaches with a special emphasis on their relevance for CAVs; and
    \item Finally, we highlight various open research problems that require further investigation.   
\end{enumerate}

%\vspace{2mm}
\textit{Organization of the Paper}: 
% \jq{Please update when the paper is finalized} 
% The rest of paper is organized as follows. 
The organization of this paper is depicted in Figure \ref{fig:outline}. The history, introduction, and various challenges associated with connected and automated vehicles (CAVs) are presented in Section \ref{sec:back}. Section \ref{sec:MLpipeline} presents an overview of the ML pipeline in CAVs. The detailed overview of adversarial ML and its threats for CAVs are described in Section \ref{sec:adv_ML_CAVs}. An outline of various solutions to robustify applications of ML along with common methods and recommendations for evaluating robustness are presented in Section \ref{sec:robust_sols}. Section \ref{sec:research_issues} presents open research problems on the use of ML in the context of CAVs. Finally, we conclude the paper in Section \ref{sec:con}. A summary of the salient acronyms used in this paper is presented in Table \ref{tab:acronyms} for convenience.

\begin{table}[!ht]
\centering
\scriptsize
%\tiny
\caption{List of Acronyms}
\begin{tabular}{|l|l|}
\hline 
BSM & Basic Safety Message \\\hline
BFGS & Broyden--Fletcher--Goldfarb--Shanno Algorithm\\\hline
CAN & Controller Area Network \\\hline
CAVs & Connected and Automated (Autonomous) Vehicles\\\hline
CIFAR & Canadian Institute for Advanced Research \\\hline
CNN & Convolutional Neural Network \\\hline
C\&W & Carlini and Wagner Algorithm \\\hline
DARPA & Defense Advanced Research Projects Agency \\\hline
DL & Deep Learning \\\hline
DNN & Deep Neural Network \\\hline
ECUs & Electronic Control Units \\\hline
FGSM & Fast Gradient Sign Method \\\hline
GAN & Generative Adversarial Networks \\\hline
GPS & Global Positioning System \\\hline
GTSDB & German Traffic Sign Detection Benchmark\\\hline
GTSRB & German Traffic Sign Recognition Benchmark\\\hline
IoV & Internet of Vehicles \\\hline
JSMA & Jacobian-based Saliency Map Attack \\\hline
L-BFGS & Limited-memory BFGS\\\hline
LIDAR & LIght Detection and Ranging\\\hline
LISA & Laboratory for Intelligent \& Safe Automobiles \\\hline
LSTM & Long Short-Term Memory \\\hline
MC/DC & Modified Condition/Decision Coverage\\\hline
ML & Machine Learning \\\hline
MNIST & Modified National Institute of Standards and Technology \\\hline
ODD & Operational Design Domain\\\hline
RADAR & RAdio Detection And Ranging \\\hline
RL & Reinforcement Learning \\\hline
RSU & Road-Side Unit \\\hline
SAE & Society of Automotive Engineers\\\hline
SVM & Support Vector Machine \\\hline
VANETs & Vehicular Ad-hoc Networks \\\hline
V2I & Vehicle to Infrastructure \\\hline
V2V & Vehicle to Vehicle \\\hline
V2X & Vehicle to Everything \\\hline
VGG & (Oxford University's) Visual Geometry Group\\\hline
YOLO & You Only Look Once (Classifier)\\\hline
\end{tabular} 
\label{tab:acronyms}
\end{table}

\section{Connected and Autonomous Vehicles (CAVs): History, Introduction, and Challenges}
\label{sec:back}

% \jq{Refer to the following important papers for this:} \cite{petit2015potential}  \cite{parkinson2017cyber}. \jq{You can consider making tables such as those in \cite{petit2015potential}.}

%\jq{Talk of Electronic Control Units (ECUs) at an appropriate place (see the references above). Cite the attacks that can be launched on these ECUs and what damage can be caused.}

% ``Cars have become very complex computer sys­tems with about 100 million lines of code and more than 100 electronic control units (ECUs) inter­ connected to control everything, including steering, ac­celeration, brakes, and other safety­-critical systems.'' \cite{kleberger2011security}
In this section, we provide the history, introduction, background of CAVs along with different conventional and security challenges associated with them. 

\subsection{Autonomous Vehicles and Levels of Automation}

%\jq{\textit{Highlight some history of self-driving cars}:}

\begin{figure*}[!ht]
    \centering
    \includegraphics[width=0.9\textwidth]{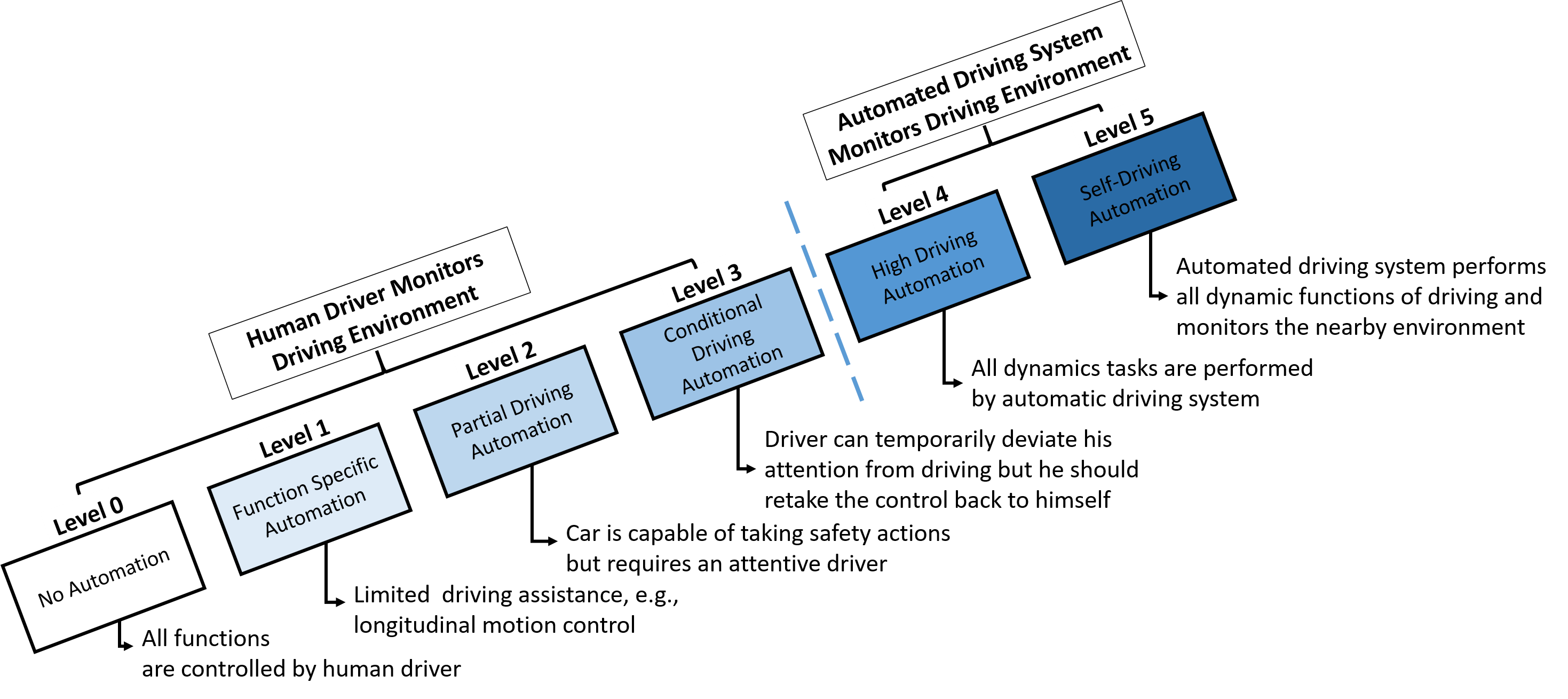}
    \caption{The taxonomy of the levels of automation in driving.}
    \label{fig:levels}
\end{figure*}

\begin{figure}[!ht]
    \centering
    \includegraphics[width=0.4\textwidth]{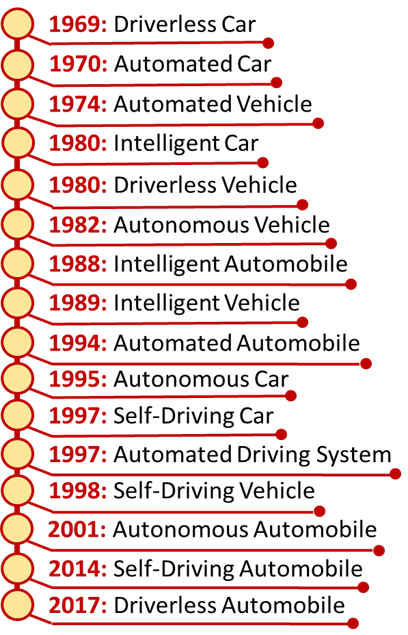}
    \caption{The illustration of different naming conventions used for referring autonomous vehicles in past years, the year denotes the publication year of first paper mentioning corresponding name. We see that self-driving car is not entirely a new concept and it is referred to through a number of terms. (Source: \cite{gandia2019autonomous})}
    \label{fig:names}
\end{figure}

The Society of Automotive Engineers (SAE) has defined a taxonomy of driving automation that is organized in six levels. The SAE defined the potential of driving automation at each level that is described next and depicted in Figure \ref{fig:levels}. Moreover, according to a recent scientometric and bibliometric review article on autonomous vehicles \cite{gandia2019autonomous}, different naming conventions have been used over the years to refer to autonomous vehicles. These names are illustrated in Figure \ref{fig:names}; note that the year denotes the publication year of first paper mentioning the corresponding name.  

\begin{itemize}
\item \textbf{Level 0} No automation: all driving tasks and major systems are controlled by a human driver; 
\item \textbf{Level 1} Function-specific automation: provides limited driver assistance, e.g., lateral or longitudinal motion control;
\item \textbf{Level 2} Partial driving automation: at least two primary control functions are combined to perform an action, e.g., lane keeping assistance and adaptive cruise control; 
\item \textbf{Level 3} Conditional driving automation: enables limited self-driving automation, i.e., allows the driver to temporarily deviate his attention from driving to perform another activity but the presence of driver is always required to retake control within a few seconds;
\item \textbf{Level 4} High driving automation: an automated driving system performs all dynamic tasks of driving, e.g., monitoring of the environment and motion control. However, the driver is capable of getting full control of the vehicle's safety-critical functions under certain scenarios; 
\item \textbf{Level 5} Self-driving automation: an automated driving system performs all dynamic functions of driving and monitors the nearby environment for the entire trip, without any human intervention at any time. 
\end{itemize}

The SAE defines the operational design domain (ODD) for the safe operation of autonomous vehicles as ``the specific conditions under which a given driving automation system or feature thereof is designed to function, including, but not limited to, driving modes'' \cite{sae2016automotive}. ODD refers to the domain of operation which an autonomous vehicle has to deal with. An ODD representing an ability to drive in good weather conditions is quite different from an ODD that embraces all kinds of weather and lighting conditions. The SAE recommends that ODD should be monitored at run-time to gauge if the autonomous vehicle is in a situation that it was designed to safely handle. 
\subsection{Development of Autonomous Vehicles: Historical Overview}

Self-driving vehicles, especially ones considering lower levels of automation (referring to the taxonomy of automation as presented in Figure \ref{fig:levels}), have existed for a long time. In 1925, Francis Udina presented a remote controlled car famously known as American wonder. In the 1939-1940 New York World's Fair, General Motors Futurama exhibited aspects of what we call self-driving car today. The first work around the design and development of self-driving vehicles was initiated by General Motors and RCA in early 1950 \cite{shladover1990roadway} that was followed by Prof. Robert Fenton at The Ohio State University from 1964--80.

In 1986, Ernst Dickens at University of Munich designed a robotic van that can drive autonomously without traffic and by 1987 the robotic van drove up to 60 Km\/hr. This group had also started the development of video image processing to recognize driving scenes \cite{dickmanns2002vision} and it was followed by a demonstration performed under Eureka Prometheus project. The super-smart vehicle systems (SSVS) program in Europe \cite{glathe1994prometheus} and Japan \cite{tsugawa1992super} were also based on the earlier work of Ernst Dickens. In 1992, four vehicles drove in a convoy using magnetic markers on the road for relative positioning, a similar test was repeated in 1997 with eight vehicles using radar systems and V2V communications. This work has paved the way for modern adaptive cruise control and automated emergency braking systems. This R\&D work then witnessed initiatives of programs like the PATH Program by Caltrans and the University of California in 1986, in particular, the work on self-driving got huge popularity with the demonstration of research work done by the national automated highway systems consortium (NAHSC) during 1994-98 \cite{rillings1997automated} and this climax remained until 2003.

In the year 2002, Defence Advanced Research Project Agency (DARPA) announced the grand autonomous vehicles challenge, first episode was held in 2004 where very few cars were able to navigate miles through the Mojave desert. The first grand challenge was won by Carnegie Mellons University (CMU) where their car only drove nearly seven miles where the finish line was at 140 miles. In 2005, the second episode of DARPA grand challenge was held, in this episode, five out of twenty-three teams were able to make it through to the finish line. This time Stanford University's vehicle ``Stanley'' has won the challenge. In the third episode of DARPA grand challenge in 2007, universities were invited to present the autonomous vehicles on busy roads to shift the perception of the public, tech, and automobile industries about the design and feasibility of autonomous vehicles.

In 2007, Google hired the team leads of Stanford and CMU autonomous vehicle projects and started pushing towards their self-driving car design on the public roads. By the year 2010, Google's self-driving car has navigated approximately 140 thousand miles on the roads of California in quest of achieving the target of 10 million miles by 2020. In 2013, VisLab (a spin-off company of the University of Parma) successfully completed the international autonomous driving challenge by driving two orange vans 15000 km with minimal driver interventions from University of Parma in Italy to Shanghai in China. A year later in 2014, Volvo demonstrated the road train concept where one vehicle controls several other vehicles behind it in order to avoid road congestion. In 2016, Tesla cars have started the commercial sales of highway speed intelligent cruise control based cars with minimal human intervention.

\begin{figure}[!t]
    \centering
    \includegraphics[width=0.5\textwidth]{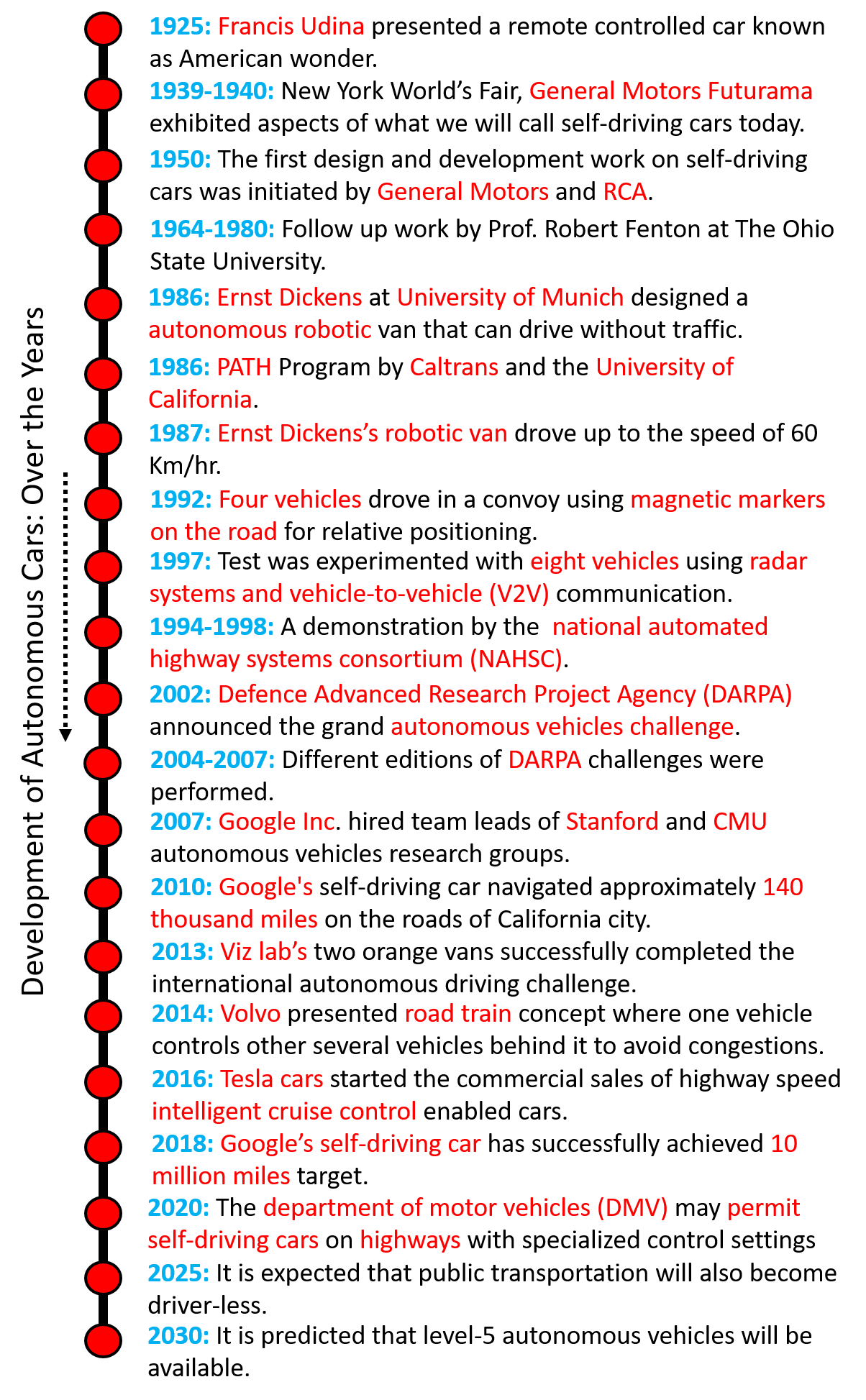}
    \caption{The timeline for the development of autonomous vehicles.}
    \label{fig:timeline}
\end{figure}

In October 2018, Google self-driving car has successfully achieved the 10 million miles target. The main aim of Google's self-driving car program is to reduce the number of deaths caused by traffic accidents by half and to date, they are working towards achieving this ambitious goal. It is expected that by 2020 the state departments of motor vehicles (DMV) may permit self-driving cars on the highways with their special lanes and control settings. By 2025, it is expected that public transportation will also become driver-less and by 2030 it is foresighted that we will have level-5 autonomous vehicles\footnote{\url{https://bit.ly/2Kei9ci}}. A timeline for the development of autonomous vehicles over the past decades is depicted in Figure \ref{fig:timeline}. 

\begin{figure}[!ht]
    \centering
    \includegraphics[width=0.5\textwidth]{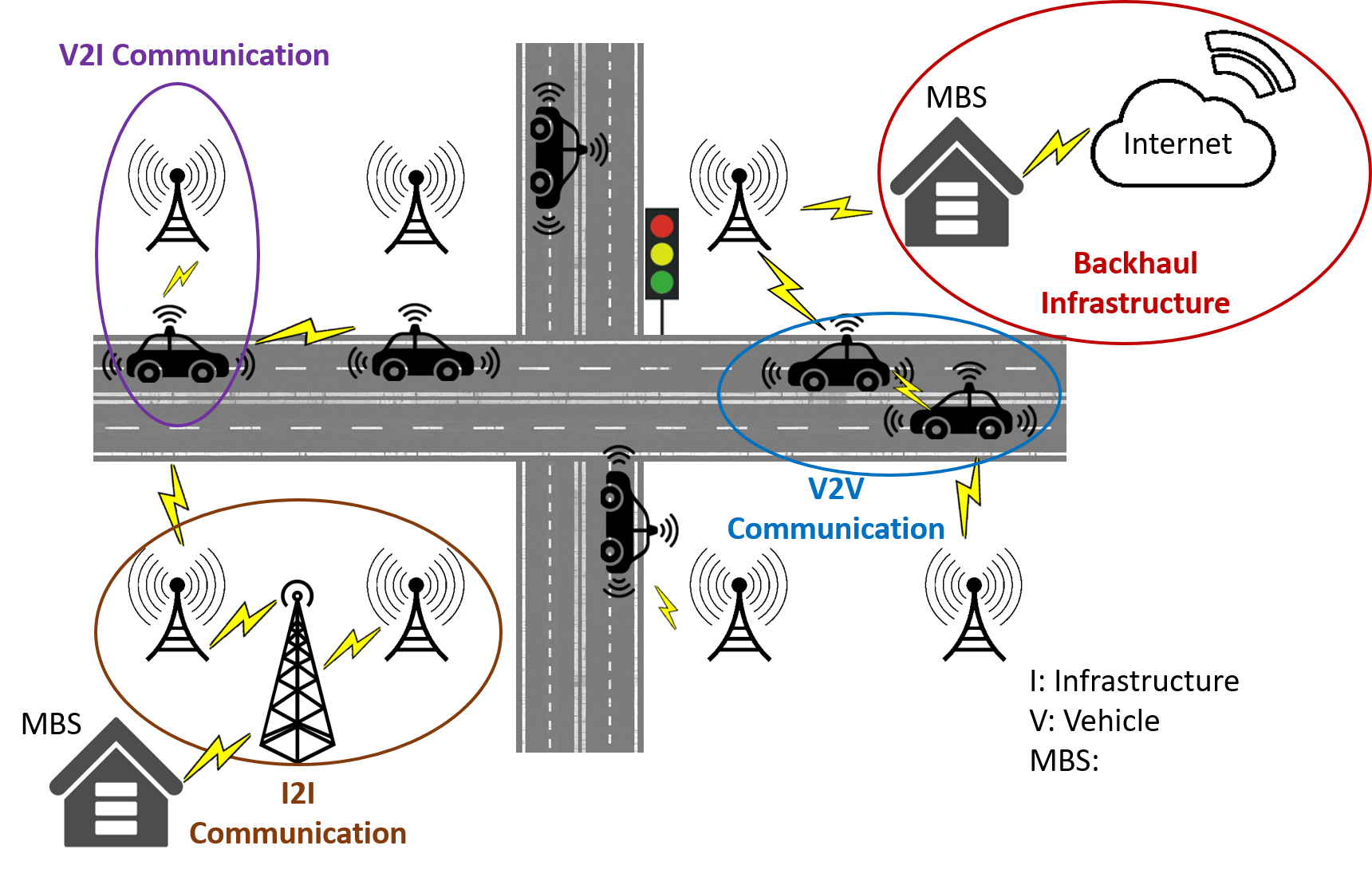}
    \caption{The basic system architecture of connected vehicles having three types of communications: vehicle-to-vehicle (V2V), infrastructure-to-infrastructure (I2I), and infrastructure-to-vehicle (I2V).}
    \label{fig:vanet}
\end{figure} 

\begin{figure*}[!ht]
    \centering
    \includegraphics[width=0.7\textwidth]{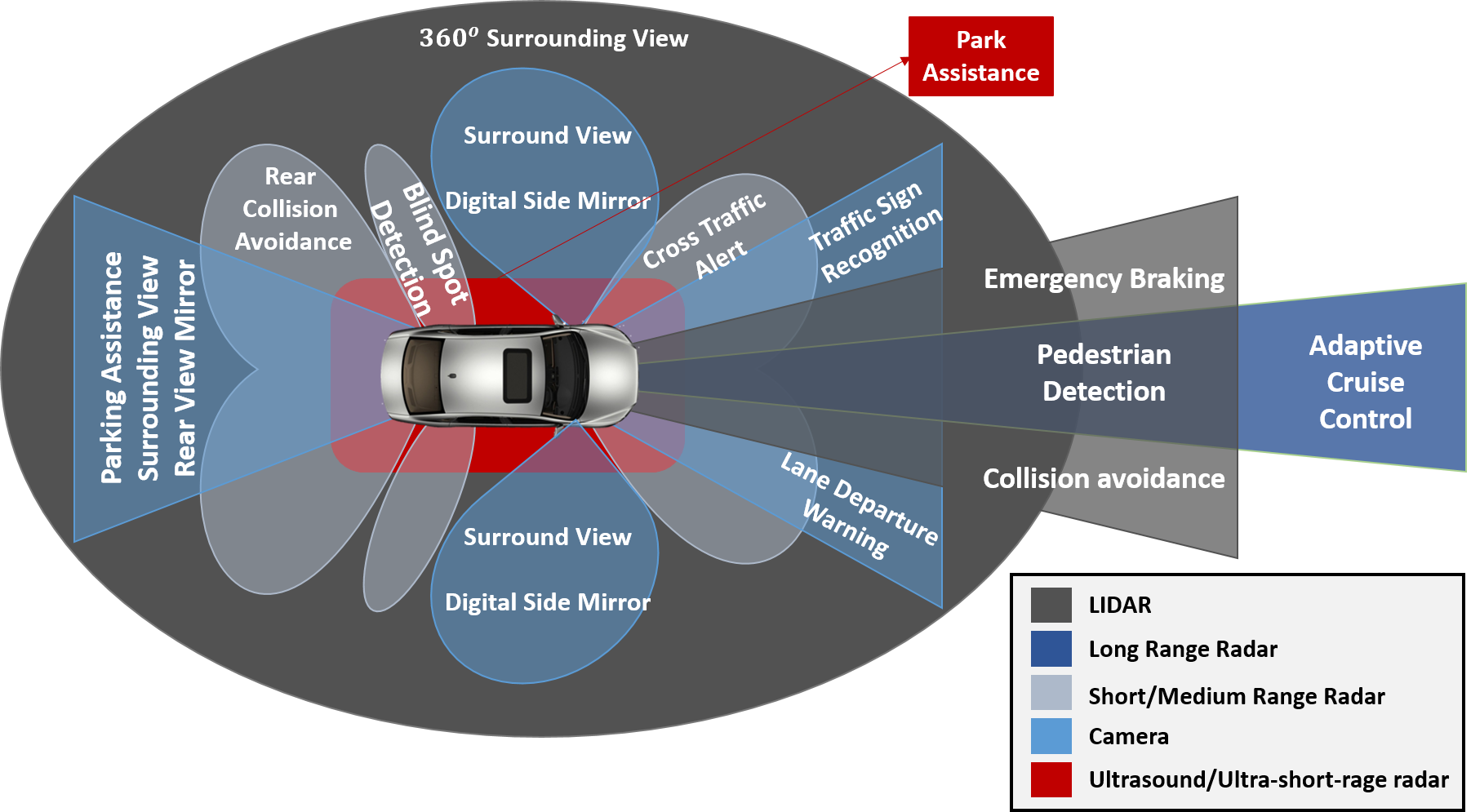}
    \caption{Autonomous vehicle's major sensor types, their range, and position (figure adapted from \cite{staszewski2013making}).}
    \label{fig:sdc}
\end{figure*}

% \jq{Highlight necessary background: such as ODD, which is important for understand automation}
% \jq{[Adapt the following] Operational Design Domain (ODD): Society for Automotive Engineers (SAE)  defines the ODD as ``The specific conditions under which a given driving automation system or feature thereof is designed to function, including, but not limited to, driving modes.'' [cite the SAE document including this text] ODD refers to the domain of operation for which an autonomous car has to deal with. An ODD representing an ability to drive in good weather conditions is quite different from an ODD that embraces all kinds of weather and lighting conditions. SAE recommends that ODD should be monitored at run-time to gauge if the ADS is in a situation that it was designed to safely handle.} 

% The challenges associated with autonomous vehicles are presented below. 

\subsection{Introduction to Connected and Autonomous Vehicles (CAVs)}

The term connected vehicles refers to the technologies, services, and applications that together enable inter-vehicles connectivity. In connected vehicles' settings, the vehicles are equipped with a wide variety of onboard sensors that communicate with each other via CAN bus and nearby communication infrastructures and vehicles (as illustrated in Figure \ref{fig:vanet}). The applications of connected vehicles include everything from traffic safety, roadside assistance, infotainment, efficiency, telematics, and remote diagnostics to autonomous vehicles and GPS.  In general, the connected vehicles can be regarded as a cooperative intelligent transport system \cite{uhlemann2015introducing} and fundamental component of the internet of vehicles (IoV) \cite{lu2014connected}. A review of truck platooning automation projects formulating the settings of connected vehicles (described earlier) together with various sensors (i.e., RADAR, LIDAR, localization, laser scanners, etc.) and computer vision techniques is presented in \cite{tsugawa2016review}. The key purpose of initiating and investigating such projects is to reduce energy consumption and personnel costs by automated operation of following vehicles. Furthermore, it has been suggested in the literature that throughput on urban roads can be doubled using vehicle platooning \cite{lioris2017platoons}.

\begin{figure*}[!h]
    \centering
    \includegraphics[width=0.9\textwidth]{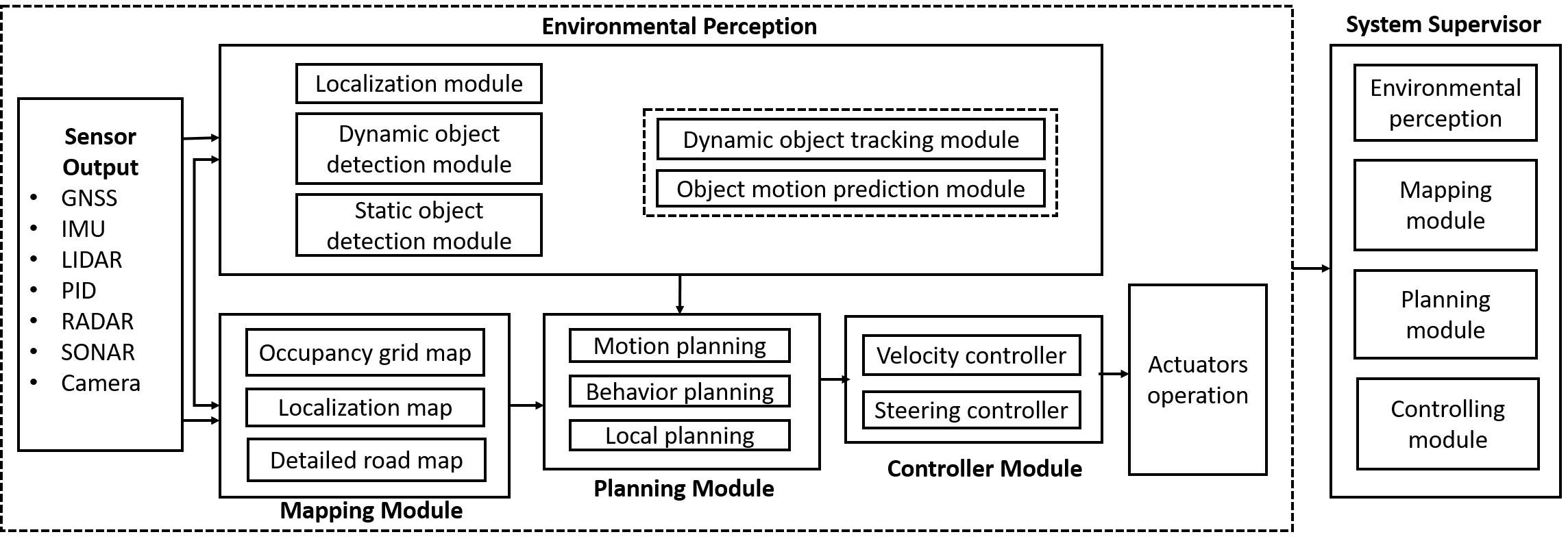}
    \caption{The systematic software workflow of autonomous vehicles. Nuts and bolts of all important operational blocks of software workflows are depicted to provide reader with a better understanding of the system design involved in developing a state-of-art autonomous vehicle.} 
    \label{fig:sdc_sys}
\end{figure*}

CAVs is an emerging area of research that is drawing substantial attention from both academia and industry. The idea of connected vehicles has been conceptualized to enable inter-vehicle communications to provide better traffic flow, road safety, and greener vehicular environment while reducing fuel consumption and travel cost. There are two types of nodes in a network of connected vehicles: (1) vehicles having onboard units (OBUs), and (2) roadside wireless communication infrastructure or roadside units (RSUs). The basic configuration of a vehicular network is shown in Figure \ref{fig:vanet}. There are three modes of communications in such networks: vehicle-to-vehicle (V2V), infrastructure-to-infrastructure (I2I), and vehicle-to-infrastructure (V2I). Besides these, there are two more types of communication---vehicle to pedestrian (V2P) and vehicle to anything (V2X)---that are expected to become part of the future connected vehicular ecosystem.

% \textit{Vehicle-to-Vehicle Communication (V2V)}: In V2V communication, vehicles exchange critical information among each others, e.g., information about accidents, road, and traffic conditions, etc. 

% \textit{Infrastructure-to-Vehicle Communication (I2V)}: In I2V communication, the infrastructure (roadside units (RSU)) broadcasts numerous messages to the moving vehicles that contain valuable information about traffic and road conditions which is used by vehicles for optimization purposes, e.g., efficient path planning. There is another similar mode of communication between the vehicle and infrastructure, i.e., vehicle-to-infrastructure (V2I) in which vehicles transmit critical safety and operational data to nearby communication infrastructures. 

% \textcolor{red}{\textit{Infrastructure-to-Infrastructure Communication (I2I)}:} 

In modern vehicles, self-contained embedded systems called electronic control units (ECUs) are used to digitally control a heterogeneous combination of components (such as brakes, lighting, entertainment, and drivetrain/powertrain, etc.) \cite{koscher2010experimental}. There are more than 100 such embedded ECUs in a car that are executing about 100 million expressions of code and are interconnected to control and provide different functionalities such as acceleration, steering, and braking \cite{kleberger2011security}. The security of ECUs can be compromised and remote attacks can be realized to gain control of the vehicle as illustrated in \cite{koscher2010experimental}.    

%\jq{Include a discussion about levels of automation in automated driving systems or autonomous vehicles}

Modern CAVs utilize a number of onboard sensors including proximity, short, middle, and long range sensors. While each of these sensors works in its dedicated range, they can act together to detect objects and obstacles over a wide range. The major types of sensors deployed in autonomous vehicles and their sensing range are shown in Figure \ref{fig:sdc} and are briefly discussed next.

\begin{itemize}
    \item \textit{Proximity Sensors (5m)}: Ultrasonic sensors are proximity sensors that are designed to detect nearby obstacles when the car is moving at a low speed, especially, they provide parking assistance.
    \item \textit{Short Range Sensors (30m)}: There are two types of short-range sensors: (1) forward and backward cameras and (2) short range radars (SRR). Forward cameras assist in traffic signs recognition and lane departure while backward cameras provide parking assistance and SRR help in blind spot detection and cross traffic alert. 
    \item \textit{Medium Range Sensors (80-160m)}: The LIDAR and medium-range radars (MRR) are designed with a medium range and are used for pedestrian detection and collision avoidance. 
    \item \textit{Long Range Sensors (250m)}: Long range radars (LRR) enable adaptive cruise control (ACC) at high speeds in conjunction with the information collected from internal sensors and from other vehicles and nearby RSU \cite{shladover2015cooperative}.
\end{itemize}

The software design of the autonomous vehicles utilizing ML/DL schemes is divided into five inter-connected modules; namely, \textit{environmental perception, a mapping module, planning module, controller module,} and \textit{system supervisor}. The software modules take input from the sensor block of autonomous vehicles and output intelligent actuator control commands. Figure \ref{fig:sdc_sys} highlights the software design of the autonomous vehicles and it also provides the sensory input required for each software module to perform the designated task. 

% \jq{There is limited discussion on the connected vehicles part of CAVs---this needs to be strengthened and particular applications of connected vehicles (such as those being studied by Usama, e.g., platooning).}

\subsection{Security-Related Challenges in Developing Robust CAVs}

%\subsection{CAVs Attack Surface}

Modern vehicles are controlled by complex distributed computer systems comprising millions of lines of code executing on tens of heterogeneous processors with rich connectivity provided by internal networks (e.g., CAN) \cite{checkoway2011comprehensive}. While this structure has offered significant efficiency, safety, and cost benefits, it has also created the opportunity for new attacks. Ensuring the integrity and security of vehicular systems is crucial, as they are intended to provide road safety and are essentially life critical. Vehicular networks are designed using a combination of different technologies and there are various attack surfaces which can be broadly classified into internal and external attacks. Different types of attacks on vehicular networks are described below.  

\subsubsection{Application Layer Attacks} The application layer attacks affect the functionality of a specific vehicular application such as beaconing and message spoofing. Application layer attacks can be broadly classified as integrity or authenticity attacks and are briefly described below.  

%The adversary can cause extreme damage to vehicular stream by injecting falsified or modified messages, data spoofing, or by launching replay attack.  
\begin{itemize}
    \item[(a)] \textit{Integrity Attacks}: In the \textit{message fabrication} attack, the adversary continuously listens to the wireless medium and upon receiving each message, fabricates its content accordingly and rebroadcasts it to the network. Modification of each message may have a different effect on the system state and depends solely on the design of the longitudinal control system. A comprehensive survey on attacks on the fundamental security goals, i.e., confidentiality, integrity, and availability can be found in \cite{sumra2015attacks}.   %\textcolor{red}{ALA: ADD REFERENCES that are specific to vehicular as this is general} 
    
    In the \textit{spoofing attack}, the adversary imitates another vehicle in the network to inject falsified messages into the target vehicle or a specific vehicle preceding the target. Therefore, the physical presence of the attacker close to the target vehicle is necessarily not required. In a recent study \cite{lim2019detecting},  the use of onboard ADAS sensors is proposed for the detection of location spoofing attack in vehicular networks. A similar type of attack in a vehicular network can be \textit{GPS spoofing/jamming attack}\cite{bittl2015emerging} in which an attacker transmits false location information by generating strong GPS signals from a satellite simulator. In addition, a thief can use integrated GPS/GSM jammer to restrain a vehicle's anti-theft system from reporting the vehicle's actual location \cite{petit2015potential}.   %\textcolor{red}{ALA: ADD REFERENCES that are specific to vehicular as this is general} 
    
    In the \textit{replay attack}, the adversary stores the message received by one of the network's nodes and tries to replay it later to attain evil goals \cite{parno2005challenges}. The replayed message contains old information that can cause different hazards to both the vehicular network and its nodes. For example, consider the message replaying attack by a malicious vehicle that is attempting to jam traffic \cite{mokhtar2015survey}. 
     %\textcolor{red}{ALA: ADD REFERENCES that are specific to vehicular as this is general} 
    
    \item[(b)] \textit{Authenticity Attacks}: Authenticity is another major challenge in vehicular networks which refers to protecting the vehicular network from inside and outside malicious vehicles (possessing falsified identity) by denying their access to the system \cite{zeadally2012vehicular}. There are two types of authenticity attacks; namely, Sybil attack and impersonation attacks \cite{li2015acpn}. In a \textit{Sybil attack}, a malicious vehicle pretends many fake identities \cite{douceur2002sybil} and in an \textit{impersonation attack}, the adversary exploits a legitimate vehicle to obtain network access and performs malicious activities. For example, a malicious vehicle can impersonate a few non-malicious vehicles to broadcast falsified messages \cite{he2015efficient}. This type of attack is also known as the masquerading attack. 
    % \textcolor{red}{ALA: ADD REFERENCES that are specific to vehicular as this is general} 
\end{itemize} 

To avoid application layer attacks, various cryptographic approaches can be effectively leveraged especially when an attacker is a malicious outsider \cite{mejri2014survey}. For instance, digital signatures can be used to ensure messages' integrity and to protect them against unauthorized use \cite{amoozadeh2015security}. In addition, digital signatures can potentially provide both data and entity level authentication. Moreover, to prevent replay attacks, a timestamp-based random number (nonce) can be embedded within messages. While the aforementioned methods are general, there are other unprecedented challenges related to vehicular networks implementation, deployment, and standardization. For example, protection against security threats becomes more challenging with the presence of a trusted compromised vehicle with a valid certificate. In such cases, data-driven anomaly detection methods can be used \cite{lyamin2018ai,ucar2017data}. A survey on anomaly detection for enhancing the security of connected vehicles is presented in \cite{rajbahadur2018survey}.
% \textcolor{red}{ALA: ADD REFERENCES that are specific to vehicular as this is general} 

\subsubsection{Network Layer Attacks}
Network layer attacks are different from the application layer attacks in a way that they can be launched in a distributed manner. One prominent example of such attacks on vehicular systems is the use of vehicular botnets to attempt a denial of service (DoS) or distributed denial of service (DDoS) attack. The potential of vehicular network-based botnet attack for autonomous vehicles is presented in \cite{garip2015congestion}. The study demonstrates that such an attack can cause severe physical congestion on hot spot road segments resulting in an increased trip duration of vehicles in the target area. Another way to realize the DoS attack is to use \textit{network jamming} that causes disruption in the communications network over a small or large geographic area. As discussed earlier, current configurations of vehicular networks are based on the IEEE 802.11p standard that uses single control channel (CCH) with multiple service channels (SCH) and can be attacked by attempting single channel or multi-channel jamming by swiping between all channels. Various conventional techniques can be adopted to mitigate network layer attacks such as frequency hopping, channel, and technology switching, etc. \textit{Coalition or platooning attack} is a similar type of attack in which a group of compromised vehicles can cooperate to perform malicious activities such as blocking or interrupting communications between legitimate vehicles. %Blackhole attack is a prime example of a coalition. 

\subsubsection{System Level Attacks}
The attacks on the vehicle's hardware and software are known as system level attacks and can be performed by either malicious insiders at the time of development or outsiders using unattended vehicular access. Such attacks are more serious in nature as they can cause damage even in the presence of the deployed state of the art security measures and secure end-to-end communications \cite{koopman2017autonomous}. For instance, if the onboard hardware or software system of a vehicle is maliciously modified then the information exchange between the vehicle and communication systems will be inaccurate and with such a phenomenon the overall performance and security of the vehicular network will be compromised. In \cite{shoukry2013non}, authors investigated a non-invasive sensor spoofing attack on car's anti-lock braking system such that the braking system mistakenly reports a specific velocity.  
% \textcolor{red}{ALA: ADD REFERENCES that are specific to vehicular as this is general; It is a good idea to also look into news articles that documented such attacks; I remember reading about an attack on the ABS braking system of a GM car, etc. Please note this comment applies to the whole section (not only this sub-section).} 

\subsubsection{Privacy Breaches}

In vehicular networks, vehicles broadcast safety messages periodically that contain critical information such as vehicle identity, current location, velocity, acceleration, etc. The adversary can exploit such kind of information by attempting an \textit{eavesdropping attack} which is a type of passive attack and is more difficult to be detected. Therefore, preserving the privacy of vehicles and drivers is of utmost importance. This allows the vehicles to communicate with each other without disclosing their identities, which is accomplished by masking their identities, e.g., using pseudonyms. In vehicular networks, knowing the origin of the message is crucial for authentication purposes, therefore, vehicles should be equipped with privacy-preserving authentication mechanism ensuring that the communication among vehicles (V2V) and with infrastructure (V2I) is confidential. However, inter-vehicular communication can be eavesdropped by anyone within the radio range, e.g., a malicious vehicle can collect and misuse confidential information. Similarly, an attacker can construct location profiles of vehicles by establishing a connection with the RSU. Therefore, the effectiveness of pseudonymous or even complete anonymous schemes in vehicular networks remains vulnerable to privacy breaches \cite{wiedersheim2010privacy}. 
% \textcolor{red}{ALA: ADD REFERENCES that are specific to vehicular as this is general; stories that document such attacks will be of interest} 

\subsubsection{Sensors Attacks} Although sensors of autonomous vehicles are by design resilient to environmental noises such as acoustical interference from nearby objects and vehicles, etc. However, current sensors cannot resist intentional noise and it can be injected to realize various attacks such as jamming and spoofing.

\subsubsection{Attacks on Perception System} The perception system of self-driving vehicles is developed using various computer vision techniques including modern ML/DL-based methods for identifying objects, e.g., pedestrians, traffics signs, and symbols, etc. The perception system of self-driving vehicle is highly vulnerable to the physical world and adversarial attacks. For example, suppose we're learning a controller $f(x)$ to predict the steering angle in an autonomous car as a function of the vision-based input (captured into a feature vector x). The adversary may introduce small manipulations (i.e., $x$ is modified into $x'$) such that the predicted steering angle $f(x')$ is maximally distant from the optimal angle $y$.

\subsubsection{Intrusion Detection}
The detection of malicious activities is one of the major challenges of VANETs. Intrusion detection systems enable the identification of various types of attacks being performed on the system, e.g., sink- and black-hole attacks, etc. Without such a system, communication in vehicular networks is highly vulnerable to numerous attacks such as selective forwarding rushing, and Sybil attacks, etc. To detect the selective forwarding attack, a trust system based method utilizing local and global detection of attacks among inter-nodes mutual monitoring and detection of abnormal driving patterns is presented in \cite{wang2016trust}. Ali et al. proposed a system for intelligent intrusion detection of gray holes and rushing attack  \cite{ali2016intelligent}.  
%\textcolor{red}{ALA: ADD REFERENCES}

% Survey on VANET security challenges and possible cryptographic solutions \cite{mejri2014survey}

\subsubsection{Certificate Revocation}

The security mechanism of vehicular networks is based on trusted certification authority (CA) that manages the identities and credentials of the vehicles by issuing valid certificates to them. The vehicles are essentially unable to operate in the system without a valid certificate and validity of certificate must be revoked after a certain amount of time. The revocation process is a challenging task administratively due to challenges such as the identification of nodes with illegitimate behavior and the need to change the registered domain. Moreover, it is necessary to restrain malicious nodes by revoking their certificates to prevent them from attacking the system. To tackle this problem, three different certificate revocation protocols have been proposed in  \cite{raya2006certificate}.

%\section{Security of Connected and Autonomous Vehicles (CAVs) and ML}
%\label{sec:VANET_ML}

% The summary of various adversarial ML attacks on autonomous vehicles is presented in Table \ref{table3}.     
% \subsubsection{Self-Driving Vehicular Networks}
 
% Why (and How) Networks Should Run Themselves \cite{feamster2017and};
% Machine learning paradigms for next-generation wireless networks \cite{jiang2017machine};
% The unreasonable effectiveness of deep learning \cite{lecun2014unreasonable}

% \vspace{2mm}

% The security of machine learning \cite{barreno2010security};
% Learning in the presence of malicious errors \cite{kearns1993learning};
% Towards evaluating the robustness of neural networks \cite{carlini2017towards};

% \vspace{2mm}
% \textit{Security and AI/IoT}:

% Artificial Intelligence and the Attack/Defense Balance \cite{schneier2018artificial}; 
% Privacy and cybersecurity: The next 100 years \cite{landwehr2012privacy}; Thinking Security: Stopping Next Year's Hackers \cite{bellovin2015thinking};   Security and the Internet of Things \cite{schneier2017security}; Outside the closed world: On using machine learning for network intrusion detection \cite{sommer2010outside}

\begin{figure*}[!ht]
    \centering
    \includegraphics[width=0.85\textwidth]{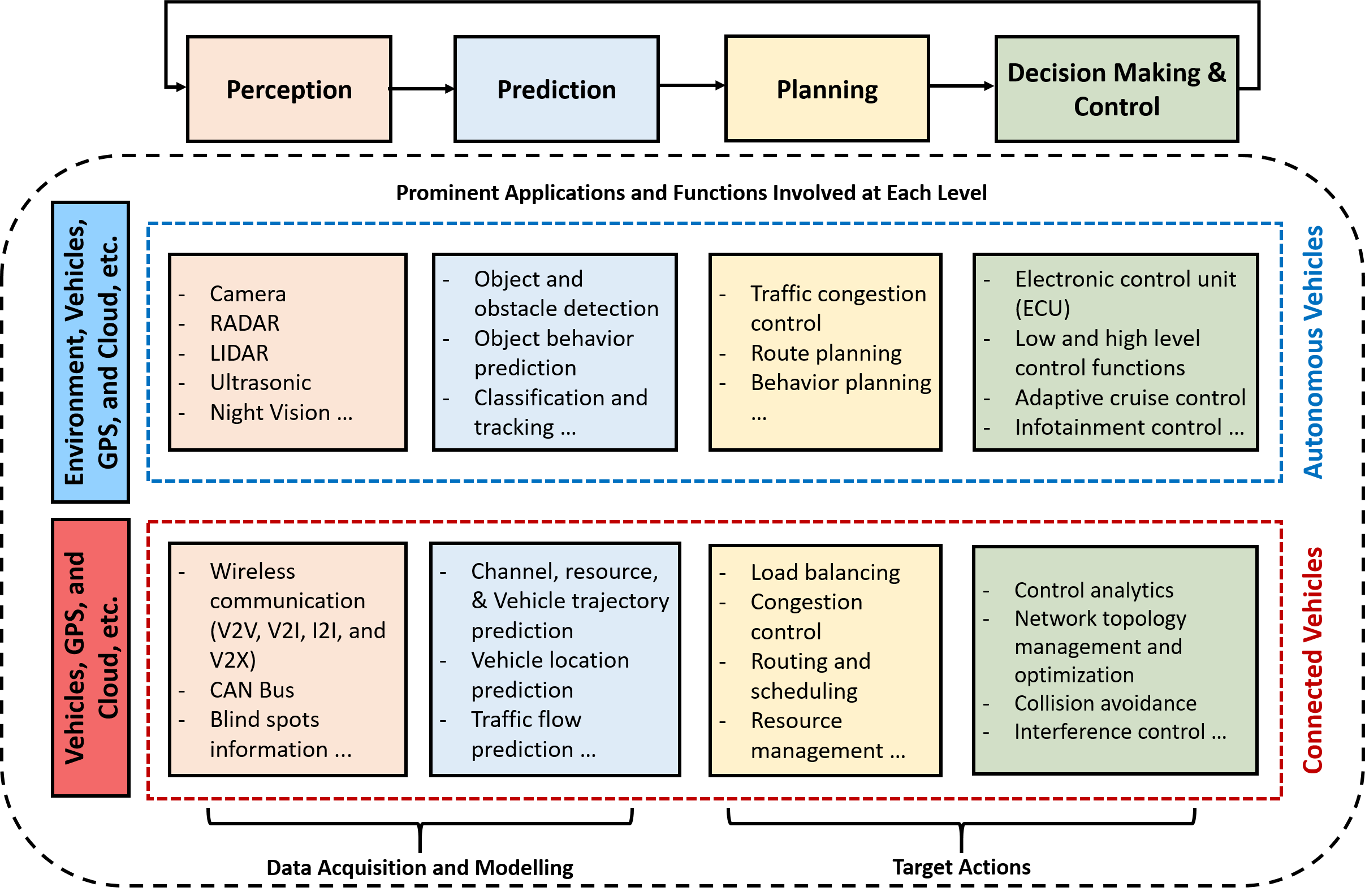}
    \caption{The \textbf{machine learning (ML) pipeline} of CAVs comprising of four major modules: (1) perception; (2) prediction; (3) planning; and (4) control.} %\jq{This is an important figure---see if you can improve it and sync it better with the rest of the paper.}} %\jq{Relate this to the ML tasks highlighted earlier of (1) Perception; (2) Prediction; (3) Planning; and (4) Control. Think of how this figure may be further enhanced.}}
    \label{fig:vanet_ml}
\end{figure*}

\subsection{Non-Security-Related Challenges in Deploying CAVs}

% \jq{Make this section to the point and short (half its current size)}

%\jq{We don't want a very elaborate discussion of traditional VANET challenges, particularly, in the short magazine article. We should highlight the major challenges succinctly and provide references to the existing papers. The bulk of the paper should be focused on the security of ML-enabled CAVs and the ML-specific challenges}.
%\jq{Discussions of traditional security challenges such as the following can come in the background section.} 
The phenomenon of connected vehicles is realized using a technology named vehicular networks which have various challenges that need to be addressed for their efficient deployment in the longer term. Various challenges associated with vehicular networks are described below.

%For instance, a major challenge is the large-scale mobility of nodes, i.e., most vehicles move continuously with relatively high speed which results in a dynamic network topology. In such a dynamic network, the links between nodes (vehicles) are very unstable and lead to various other challenges such as how to efficiently allocate resources, how to alleviate network congestion, and how to optimize the topology in such dynamic and unstable environments. Various challenges associated with vehicular networks are described below.

% The VANETs are high mobility networks that offer unprecedented challenges and exhibit distinctive characteristics as compared to other wireless networks. Various challenges associated with VANETs are described below. 
%In this section, we present such challenges and their potential solutions with a key focus on using machine learning based approaches. 

\subsubsection{High Mobility of Nodes}
% Due to the large scale mobility of vehicles, the topology of vehicular networks potentially changes quickly as compared to other wireless networks.
The large scale mobility of vehicles in vehicular networks result in a highly dynamic topology; thus, raising several challenges for the communication networks \cite{cunha2016data}. In addition, the dynamic nature of traffic can lead to a partitioned network having isolated clusters of nodes \cite{blum2004challenges}. As the connections between the vehicles and nearby RSUs are short-lived, the wireless channel coherence time is short. This makes accurate real-time channel estimation more challenging at the receiver end. This necessitates the design of dynamic and robust resource management protocols that can efficiently utilize available resources while adapting to the vehicular density variations \cite{fontes2017theory}. %For instance, a mobility-based approach for dynamic clustering of vehicles in vehicular networks that utilize different mobility metrics (e.g., vehicle density, moving direction, an relative velocity, etc.) is presented in \cite{ren2017mobility}.   

%\textcolor{red}{ALA: ADD REFERENCES}    

%As compared with low mobility networks, channel propagation characteristics are the most important discriminating factors of high mobility networks.
%VANETs suffer from high mobility of nodes (vehicles) that infers strong dynamics of the network raising several design challenges of the communication network, e.g., dynamic network topology. As compared with low mobility networks, channel propagation characteristics are the most important discriminating factors of high mobility networks.    
% \subsubsection{Unbounded Channel Access Delay}
% An other important challenge of VANETS 

\subsubsection{Heterogeneous and Stringent QoS Requirements}

In vehicular networks, there are different modes of communications that can be broadly categorized into V2V and V2I communications. In V2V communications, vehicles exchange safety-critical information (e.g., information beacons, road and traffic conditions) among each other known as basic safety messages (BSM). This communication, which can be performed periodically or when triggered by some event, requires high reliability and is sensitive to delay \cite{liang2019toward}.
% ---e.g., the maximum allowed end-to-end latency is less than 5 milliseconds and the requirement for transmission reliability is more than 99.99\% for a packet of 1600 bytes in the European METIS project\footnote{\url{https://metis2020.com/documents/deliverables/index.html}} \cite{fallgren2013scenarios}. 
In V2I communications, on the other hand, vehicles can communicate with nearby communications infrastructure to get support for route planning, traffic information, operational data, and to access entertainment services that requires more bandwidth and frequent access to the Internet, e.g., for downloading high-quality maps and accessing infotainment services, etc. Therefore, the heterogeneous and stringent QoS requirements of VANETs cannot be simultaneously met with traditional wireless design approaches. 

%To tackle this issue, high bandwidth data sharing is being considered in the 3GPP project to support V2X communication with the evolution of future 5G cellular networks. 

%Such communication is considerably more bandwidth intensive as it requires frequent access to the Internet for downloading high-quality maps, media streaming, and to access Internet-enabled services such as social networking. The heterogeneous and stringent QoS requirements of VANETs cannot be simultaneously met with traditional wireless design approaches and high bandwidth data sharing is being considered in the 3GPP project to support V2X communication with the evolution of future 5G cellular networks.

% Much research has focused on addressing this issue---e.g., high bandwidth data sharing is being considered in the 3GPP project to support V2X communication with the evolution of future 5G cellular networks. The purpose of developing V2X technology is to enable the communication between all entities encountered in the road environment including vehicles, communications infrastructure, pedestrians, cycles, etc.  

%\subsubsection{Partitioned Network}

% \jq{Also see this paper: Securing the connected car: a security-enhancement methodology \cite{strandberg2018securing}}
  
\subsubsection{Learning Dynamics of Vehicular Networks}
% This leads to many challenges such as dynamic network topology, dynamic traffic, etc. 
As discussed above, vehicular networks exhibit high dynamicity; thus, to meet the real-time and stringent requirements of vehicular networks, historical data-driven predictive strategies can be adopted, e.g., traditional methods like hidden Markov models (HMM) and Bayesian methods \cite{liang2019toward}. In addition to using traditional ML methods, more sophisticated DL models can be used, for example, recurrent neural networks (RNN) and long short term memory (LSTM) have been shown beneficial for time series data and can be potentially used for modeling temporal dynamics of vehicular networks. 
% However, the security of ML/DL models is itself an open research problem which we discuss in later sections. %\textcolor{red}{ALA: ADD REFERENCES}

\subsubsection{Network Congestion Control}
Vehicular networks are geographically unbounded and can be developed for a city, several cities, and countries as well. The unbounded nature of vehicular networks leads to the challenge of network congestion \cite{ye2018deep}. As the traffic density is high in urban areas as compared to rural areas, particularly during rush hours, that can possibly lead to network congestion issues.  %\textcolor{red}{ALA: ADD REFERENCE}

\subsubsection{Time Constraints} 
The efficient application of vehicular networks requires hard real-time guarantees because it lays out the foundation for many other applications and services that require strict deadlines \cite{yang2004vehicle}, for example, traffic flow prediction \cite{wan2016mobile}, traffic congestion control \cite{de2016real}, and path planning \cite{wang2015real}. Therefore, safety messages should be broadcasted in acceptable time either by vehicles or RSUs.

\begin{table*}[!t]
\centering
\caption{Overview of machine learning (ML)-based research on different vehicular network's applications}
\begin{tabular}{|l|l|l|}
\hline
\multicolumn{1}{|c|}{Authors} & \multicolumn{1}{c|}{Application}                                  & \multicolumn{1}{c|}{Methodology} \\ \hline
Yao et al. \cite{yao2018v2x}                    & \multirow{3}{*}{Location prediction based scheduling and routing} & Hidden Markov models             \\ \cline{1-1} \cline{3-3} 
Xue et al. \cite{xue2012novel}                   &                                                                   & Variable-order Markov models     \\ \cline{1-1} \cline{3-3} 
Zeng et al. \cite{zeng2017channel}                  &                                                                   & Recursive least squares          \\ \hline
Karami et al. \cite{karami2015accpndn}                & \multirow{2}{*}{Network congestion control}                       & Feed forward neural network      \\ \cline{1-1} \cline{3-3} 
Taherkhani et al. \cite{taherkhani2016centralized}            &                                                                   & k-means clustering               \\ \hline
Li et al.  \cite{li2017user}                    & Load balancing                                                    & Reinforcement learning           \\ \hline
Taylor et al. \cite{taylor2016anomaly}                 & Network security                                                  & LSTM                             \\ \hline
Zheng et al. \cite{zheng2016delay}                  & Virtual resource allocation                                       & Reinforcement learning           \\ \hline
Atallah et al. \cite{atallah2017reinforcement,atallah2017deep}                & Resource management                                               & Reinforcement learning           \\ \hline
Ye et al. \cite{ye2018deep}                     & Distributed resource management                                   & Reinforcement learning           \\ \hline
Kim et al. \cite{kim2017probabilistic}                   & Vehicle trajectory prediction                                     & Reinforcement learning           \\ \hline
\end{tabular}
\label{tab:ML}
\end{table*}

\section{The ML Pipeline in CAVs}
\label{sec:MLpipeline}

%\jq{I have broken down the background section into two sections. Please start/introduce this section properly.}
%\subsection{ML-based Solutions for CAVs}

The driving task elements of self-driving vehicles that can benefit from ML can be broadly categorized into the following four major components (as shown in Figure \ref{fig:vanet_ml}). 
\begin{enumerate}
\item \textbf{Perception}: assists in perceiving the nearby environment and recognizing objects;
\item \textbf{Prediction}: predicting the actions of perceived objects, i.e., how environmental actors such as vehicles and pedestrians will move;
\item \textbf{Planning}: route planning of vehicle, i.e., how to reach from point A to B; 
\item \textbf{Decision Making \& Control}: making decisions relating to vehicle movement, i.e., how to make the longitudinal and lateral decisions to control and steer the vehicle.
\end{enumerate}

These components are combined to develop a feedback system for enabling the phenomenon of self-driving without any human intervention. This ML pipeline can then facilitate autonomous real-time decisions by leveraging insights from the diverse types of data (e.g., vehicles' behavioral patterns, network topology, vehicles' locations, and kinetics information, etc.) that can be in easily gathered by CAVs. 

In the remainder of this section, we will discuss some of the most prominent applications of ML-based methods for performing these tasks (a summary is presented in Table \ref{tab:ML}).

%(e.g., such as vehicle trajectory and traffic flow prediction, etc. CAVs can use ML to leverage  This can allow the learning of the dynamics of connected vehicles, which can be used to optimize the network (e.g., through efficient resource allocation and routing). 
%In this section, we provide a broad overview of the ML pipeline used in CAVs.
%We start by first discussing the potential of using ML for CAVs' applications. 
%\jq{There is mismatch in the discussion in the section and in the figure. Please reconcile.}
%In modern CAVs, ML and DL techniques have been increasingly deployed for a range of applications and the most prominent applications of ML-based methods in modern CAVs are described below.  

\subsection{Applications of ML for the Perception Task in CAVs}

Different ML techniques, particularly, DL models have widely been used for developing the perception system of autonomous vehicles \cite{yan2018supervised}. In addition to using video cameras as major visionary sensors, these vehicles also use other sensors for detection of different events in the car's surroundings, e.g., RADAR and LIDAR. The surrounding environment of the autonomous vehicles is perceived in two stages \cite{rosique2019systematic}. In the first stage, the whole road is scanned for the detection of changes in the driving conditions such as traffic signs and lights, pedestrian crossing, and other obstacles, etc. In the second stage, knowledge about the other vehicles is acquired. In \cite{Chen_2015_ICCV}, a CNN model is trained for developing direct perception representation of autonomous vehicles.      

\subsection{Applications of ML for the Prediction Task in CAVs}
% \jq{Examples of Prediction are missing.}
In CAVs, accurate and timely prediction of different events encountered in driving scenes is another important task which is mainly accomplished using different ML and DL algorithms. For instance, autonomous vehicles uses DL models for the detection and localization of obstacles \cite{ramos2017detecting}, different objects (e.g., vehicles, pedestrians, and bikes, etc.) \cite{chavez2016multiple} and their behavior (e.g., tracking pedestrians along the way \cite{wang2017pedestrian}) and traffic signs \cite{zeng2017traffic} and traffic lights recognition \cite{john2014traffic}. Another prediction tasks in CAVs that involve the application of ML/DL methods are vehicle trajectory and location prediction \cite{lin2018deep}, efficient and intelligent wireless communication \cite{mao2018deep}, and traffic flow prediction and modeling \cite{ye2018modeling}. Moreover, ML schemes have also been used for the prediction of uncertainties in autonomous driving conditions \cite{zhang2019neural}. 

\subsection{Applications of ML for the Planning Task in CAVs}
CAVs are equipped with onboard data processing compatibilities and they intelligently process the data collected from heterogeneous sensors for efficient route planning and for other optimized operations using different ML and DL techniques. The key goal of route planning is to reach the destination in a small amount of time while avoiding traffic congestion, potholes, and other vehicles by navigating through GPS and consuming less fuel as possible. In the literature, motion planning of autonomous vehicles is studied in three dimensions: (1) finding a path for reaching destination point; (2) searching for the fastest manoeuvre; and (3) determining the most feasible trajectory \cite{katrakazas2015real}. Moreover, to avoid collisions between vehicles in CAVs, predicting the trajectories of other vehicles is a crucial task for the planning trajectory of an autonomous vehicle \cite{houenou2013vehicle}. For instance, Li et al. presented a hybrid approach to model uncertainty in vehicle trajectory prediction for CAVs application using deep learning and kernel density estimation \cite{li2018modeling}.

\begin{figure*}[!ht]
    \centering
    \includegraphics[width=0.65\textwidth]{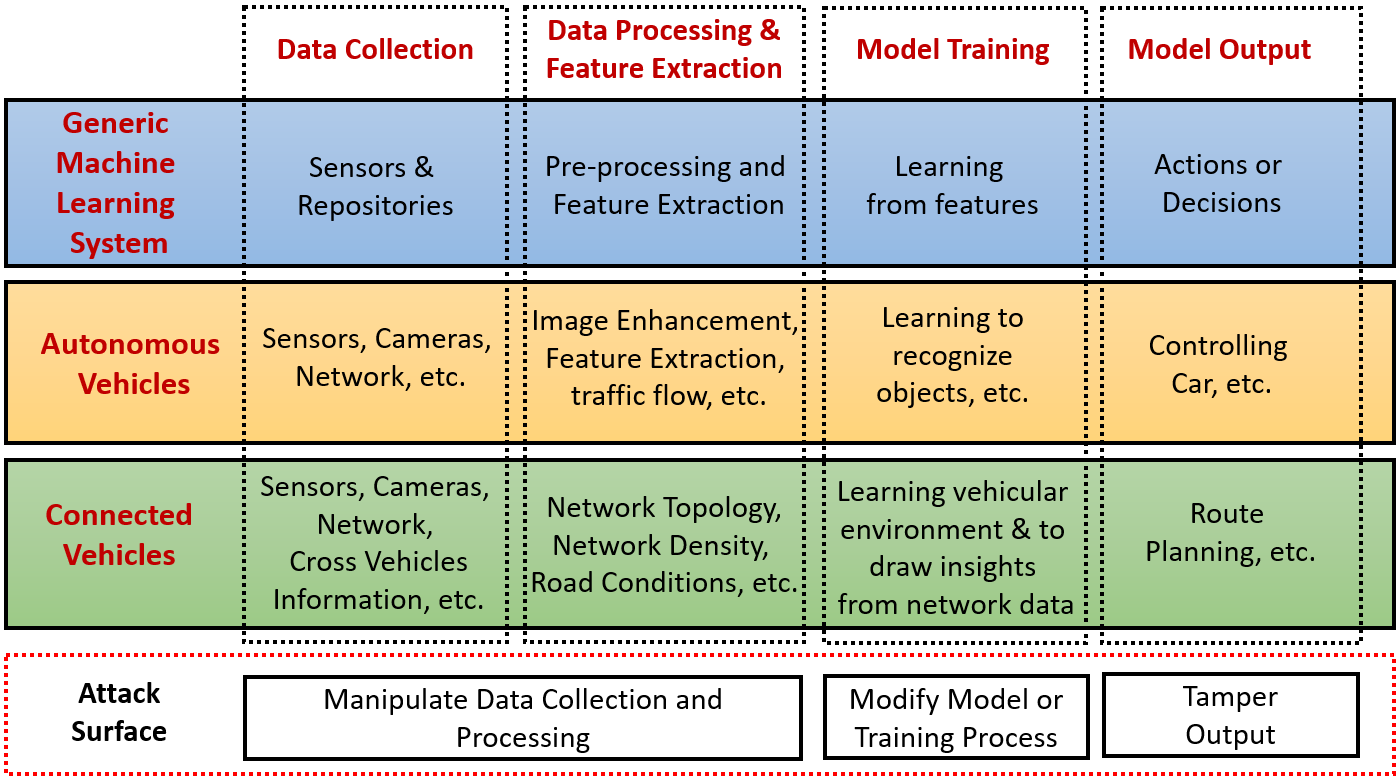}
    \caption{The illustration of the generalization of attack surfaces in ML systems: generic model (top), autonomous vehicles model (middle), and connected vehicles model (bottom).} 
    \label{fig:sdc_attack_surface}
\end{figure*}

\subsection{Applications of ML for the Decision Making and Control}
In recent years, DL based algorithms have been extensively used for control of autonomous vehicles that are refined through millions of kilometers of test drives. For instance, Bojarski et al. presented a CNN based end-to-end learning framework for self-driving cars \cite{bojarski2016end}. The model was able to drive the car on local roads with or without markings and on highways with small training data. In a similar study, CNN is trained for end-to-end learning of lane keeping for autonomous cars \cite{chen2017end}. Recently, researchers have now started working on utilizing deep reinforcement learning (RL) for performing actions and decision making in driving conditions \cite{liu2018elements}. Bouton et al. proposed a generic approach to enforce probabilistic guarantees on RL learning for which they derived an exploration strategy that restricts the RL agent to choose among only those actions that satisfy a desired probabilistic specification criteria prior to training \cite{bouton2019reinforcement}. Moreover, human-like speed control of autonomous vehicles using deep RL with double Q-learning is presented in \cite{zhang2018human} that uses scenes generated by naturalistic driving data for learning. In \cite{he2018integrated}, authors presented an integrated framework that uses a deep RL based approach for dynamic orchestration of networking, caching, and computing resources for connected vehicles.

In addition, ML-based methods have been used for many other applications in CAVs. For example, \textit{adaptive traffic flow} in which smart infrastructure integrates V2V signals from the moving cars to optimize speed limits, traffic-light timing, and the number of lanes in each direction on the basis of the actual traffic load. The traffic flow can be further improved in CAVs by using \textit{cooperative adaptive cruise control} technology \cite{milanes2014cooperative}. Also, vehicles can take advantage of cruise control and save fuel by following one another in the form of vehicles platoons. Moreover, DL based methods have been proposed for intrusion detection for in-vehicle security of CAN bus \cite{kang2016intrusion}. The overview of intelligent and connected vehicles, current and future perspectives are presented in \cite{yang2018intelligent}.  

% In modern CAVs, ML and DL techniques have been increasingly deployed from perception \cite{yan2018supervised}, planning and decision making \cite{schwarting2018planning} to intelligent traffic monitoring \cite{pawlowicz2018smart}. Li et al. \cite{li2018modeling} presented a hybrid approach to model uncertainty in vehicle trajectory prediction for CAVs application using deep learning and kernel density estimation. In \cite{he2018integrated}, authors presented an integrated framework that uses a deep reinforcement learning approach for dynamic orchestration of networking, caching, and computing resources for connected vehicles. Moreover, DL based methods have been proposed for intrusion detection for in-vehicle security of CAN bus \cite{kang2016intrusion}. The overview of intelligent and connected vehicles, current and future perspectives are presented in \cite{yang2018intelligent}. 

Autonomous vehicles are evolving through four stages of development. The first stage includes passive warning and convenience systems such as front and backward facing cameras, cross-traffic warning mechanism, radar for blind spot detection, etc. These warning systems use different computer vision and machine learning techniques to perceive the surrounding views of the vehicle on the road and to recognize traffic signs, static, and moving objects. In the second stage, these systems are used to assist the active control system of the vehicle while parking, braking, and to prevent backing over unseen objects. In the third stage, the vehicle is equipped with some semi-autonomous operations---as the vehicle may behave unexpectedly and the on seat driver should be able to resume control. In the final stage, the vehicle is designed to perform fully autonomous operations.   

CAVs together formulate the settings of the self-driving vehicular network and there is a strong synergy between them \cite{shladover2018connected}. In addition, autonomous vehicles are an important component of future vehicular networks that are equipped with complex sensory equipment.
The autonomous vehicular networks are predictive and adaptive to their environments and are designed with two fundamental goals, i.e., autonomy and interactivity. The first goal enables the network to monitor, plan, and control itself and the later ensures that the infrastructure is transparent and friendly to interact with.

The deployment of ML in CAVs entails the following stages: 

\begin{enumerate}
    \item[(a)] \textit{Data Collection}: Input data is collected using sensors or from other digital repositories. In autonomous vehicles, input data is collected using a complex sensory network, e.g., cameras, RADAR, GPS, etc. (see Figure \ref{fig:sdc}); in a connected vehicular ecosystem, there is also inter-vehicle information communication.
    
    \item[(b)] \textit{(Pre-)Processing}: The heterogeneous data (video imagery, network, and traffic information, etc.) collected by the sensors is then digitally processed and appropriate features (e.g., traffic signs information and traffic flow information, etc.) are extracted.
    
    \item[(c)] \textit{Model Training}: Using the extracted features from the input data, a ML model is trained to recognize and distinguish between different objects events encountered in the driving environment, e.g., recognizing moving objects like pedestrian, vehicles, and cyclists, etc. and distinguishing between traffic signs, i.e., stop or speed limit sign, etc. 
    
    \item[(d)] \textit{Decision or Action}: A decision or an action  (e.g., stopping the car at the stop sign and predicting traffic flow based on the knowledge acquired by the vehicular network) is performed according to the learned knowledge and underlying system. 
\end{enumerate}

\begin{figure*}[!ht]
    \centering
    \includegraphics[width=0.8\textwidth]{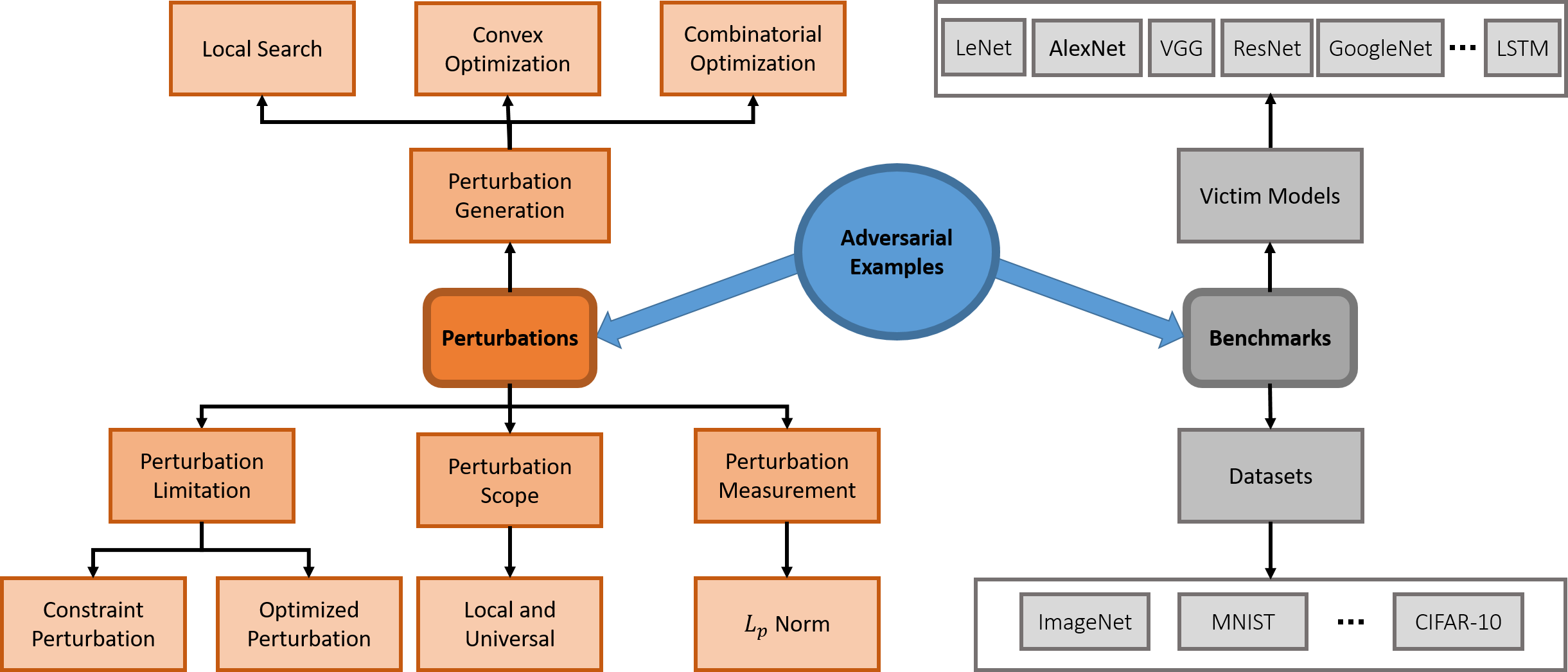}
    \caption{The taxonomy of adversarial examples, perturbation methods, and benchmarks (datasets and models).} 
    \label{fig:adv_ex_tax}
\end{figure*}

We present an illustration of the generalization of attack surfaces in ML systems from generic models to the more specific cases of autonomous and connected vehicles in Figure \ref{fig:sdc_attack_surface}. As we shall discuss later in the paper, each of these stages is vulnerable to adversarial intrusion since an adversary can try to manipulate the data collection and processing system, tamper the model, or its outputs.

\section{Adversarial ML Attacks and the Adversarial ML Threat for CAVs}
\label{sec:adv_ML_CAVs}
A comprehensive overview of adversarial ML in the context of CAVs is presented in this section. 
%In this section, we provide an overview of Adversarial ML and CAVs' security.
%where we describe various attacks surfaces and different threat models mainly in the context of security of ML. 

\subsection{Adversarial Examples}

Formally, adversarial examples are defined as inputs to a deployed ML/DL model created by an attacker by adding an imperceptible perturbation in the actual input to compromise the integrity of the ML/DL model. An adversarial sample $x^*$ is created by adding a small carefully crafted perturbation $\delta$ to the correctly classified sample $x$. The perturbation $\delta$ is calculated by approximating the optimization problem given in Eq. \ref{equ1} iteratively until the crafted adversarial example gets classified by ML classifier $f(.)$ in targeted class $t$. A taxonomy of adversarial examples, perturbation methods, and benchmarks is presented in Figure \ref{fig:adv_ex_tax}.

\begin{equation}
    x^* = x + \arg \underset{\delta{_x}}{\text{min}} \{\|\delta\|: f(x + \delta) = t\} %\underset{S}{\text{min}}\min{_\delta}
    \label{equ1}
\end{equation}

\subsubsection{Adversarial Attacks}
An adversarial attack affecting the training phase of the learning process is termed as poisoning attack where an attacker compromises the learning process of the ML/DL scheme by manipulating the training data \cite{biggio2012poisoning}, whereas the adversarial attack on the inference phase of the learning process is termed as evasion attack where an attacker manipulates the test data or real-time inputs to the deployed model for producing a false result \cite{biggio2013evasion}. Usually, examples used for fooling the ML/DL schemes at inference time are called adversarial examples.

\subsubsection{Adversarial Perturbations}
The adversarial perturbation crafting is divided in three major categories; namely, \textit{local search, combinatorial optimization,} and \textit{convex relaxation}. This division is based on solving the objective function given in Eq. \ref{equ1}. Local search is the most common method of generating adversarial perturbations where the adversarial examples are generated by solving the objective function provided in Eq. \ref{equ1} to obtain a lower bound on the adversarial perturbation by using gradient-based methods. A prime example of local search adversarial example crafting is the \textit{fast gradient sign method} (FGSM) where an adversarial example is created by taking a step in the direction of the gradient \cite{goodfellow2014explaining}. In another study, the authors demonstrated that adversarial images are very easy to be constructed using evolutionary algorithms or gradient ascent \cite{nguyen2015deep}.
Combinatorial optimization is also a method for creating adversarial examples where we find the exact solution of the optimization problem provided in Eq. \ref{equ1}, a major shortcoming of this method is the increase in the computational complexity with the increase of the number of examples in the dataset. Recently, Khalil et al. \cite{khalil2018combinatorial} launched a successful adversarial attack based on combinatorial and integer programming on binarized neural networks but the performance of the proposed attack reduces as the size and dimensions of data increase. Recently, convex relaxation is also used to generate \cite{balda2018generation} and defend \cite{wong2018provable} against adversarial examples where the upper bound on the objective function provided in Eq. \ref{equ1} is calculated.   

\subsubsection{Different Aspects of Perturbations}

The adversarial examples are designed to look like the original ones and imperceptible to humans. In this regard, the addition of small perturbations is of utmost importance. Whereas, the literature suggests that even one-pixel perturbation is often sufficient to fool the deep model trained for classification task \cite{su2019one}. Here we analyze different aspects of adversarial perturbations. 
\begin{itemize}
    \item[(a)] \textit{Perturbation Scope}: Adversarial perturbations are generated from two aspects: (1) perturbations for each legitimate input and (2) universal perturbations for the complete datasets, i.e., for each original cleaned sample. To date, most of the studies considered the first scope of adversarial perturbations.  
    \item[(b)] \textit{Perturbation Limitation}: Similarly, there are two types of limitations, optimizing the system at a low perturbations scale and optimizing the system at a low perturbations scale with constrained optimization. 
    \item[(c)] The magnitude of the perturbations is mainly measured using three norms $L_2$, $L_\infty$, and $L_0$ norm. In $L_2$-norm-based attacks, the attacker aims to minimize the squared error between the original and adversarial example. $L_2$-norm measures the Euclidean distance between the adversarial example and the original sample and results in a very small amount of noise added to the adversarial sample. $L_\infty$ attacks are perhaps the simplest type of attacks which aim to limit or minimize the extent to which the maximum change for all pixels in adversarial examples is achieved. Also, this constraint forces to only make very small changes to each pixel. $L_0$-norm-based attacks work by minimizing the number of perturbed pixels in an image and force the modifications only to very few pixels.
\end{itemize}

\begin{figure*}[!ht]
    \centering
    \includegraphics[width=0.7\textwidth]{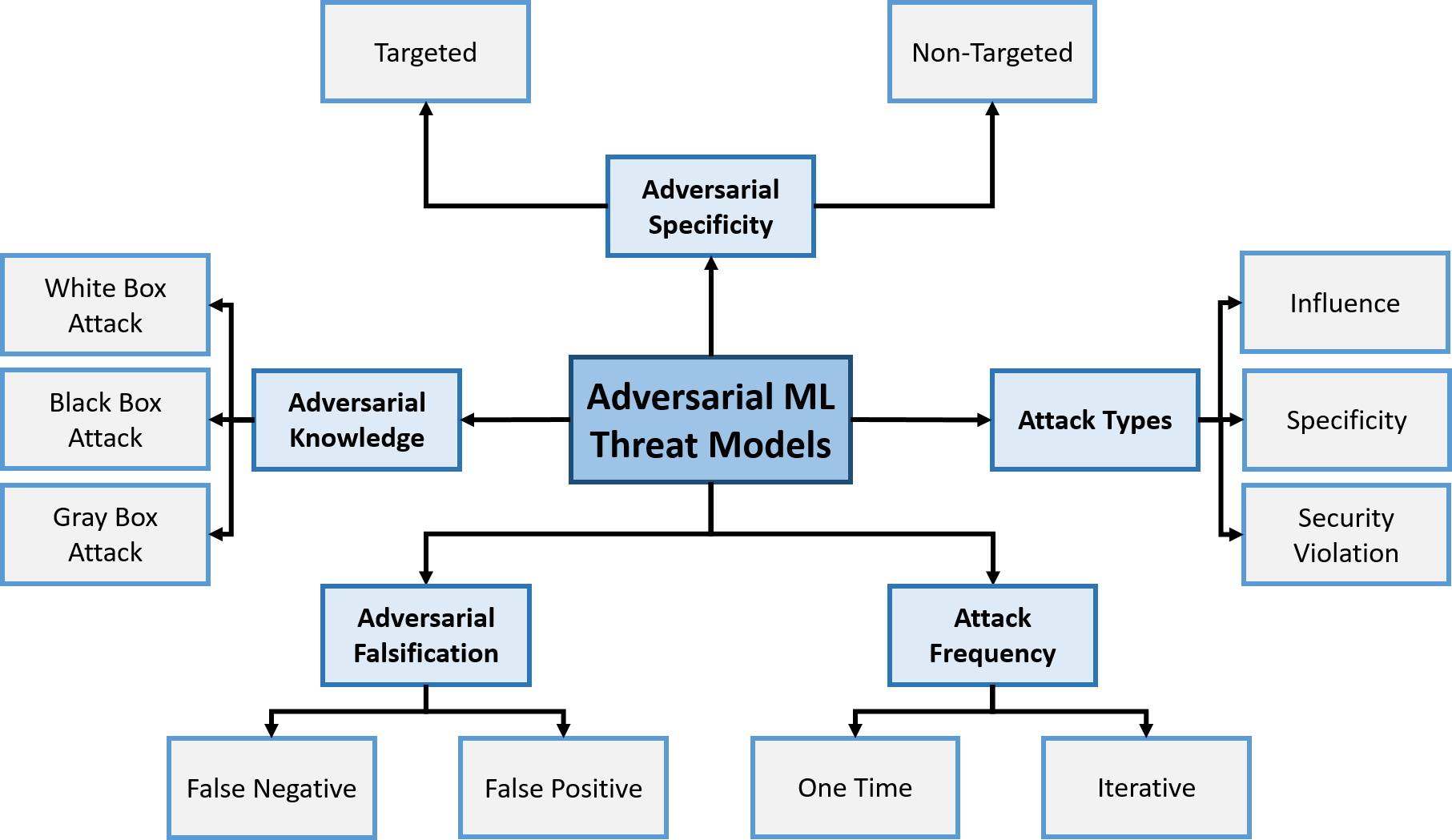}
    \caption{The taxonomy of various types of threat models used in literature to design adversarial ML attacks. This figure also provides the information needed by a defender to ensure the robustness of ML-based autonomous system.} 
    \label{fig:th_models}
\end{figure*}

To ensure tightly constrained action space available to an adversary, imperceptibility of perturbations is important to develop an attack. Considering the important constraints: (1) what constraints are placed on the attacker's ``starting point''? and (2) where did this
initial example come from? Gilmer et al. identified four salient features (described below) of adversarial perturbations \cite{gilmer2018motivating}.

\begin{itemize}
    \item[(a)] \textit{Indistinguishable Perturbation}: The attacker does not have to select a starting point but it is given a draw from the data distribution and introduces such perturbation in the input sample that is indistinguishable by a human. 
    \item[(b)] \textit{Content-Preserving Perturbation}: The attacker does not have to select a starting point but it is given a draw from the data distribution and creates such perturbation as long as the original content of the sample is preserved.   
    \item[(c)] \textit{Non-suspicious Input}: The attacker can generate any type of desired perturbed input sample as long as it remains undetectable to a human. 
    \item[(d)] \textit{Content-Constrained Input}: The attacker can generate any type of desired perturbed input sample as long as it maintains some content payload, i.e., it must be a picture of dog but not necessarily a particular dog. This includes payload-constrained input, where human perception might not be important. Rather, the intended function of the input example remains intact. 
    \item[(e)] \textit{Unconstrained Input}: There is no constraint on the input and an attacker can produce any type of input example to get the desired output or behavior from the system. 
\end{itemize}

\subsubsection{Adversarial ML Benchmarks}
In this section, we describe the benchmarks datasets and victim ML models used for evaluating adversarial examples. Researchers mostly adopt an inconsistent approach and report the performance of the attacks on diverse datasets and victim models. The widely used benchmark datasets and victim models are described below.  
\begin{itemize}
    \item \textit{Datasets}: MNIST \cite{lecun1998gradient}, CIFAR-10 \cite{krizhevsky2009learning}, and ImageNet \cite{deng2009imagenet} are the widely used datasets in adversarial ML research and are also regarded as the standard deep learning datasets. 
    \item \textit{Victim Models}: The widely used victim ML/DL models for evaluating adversarial examples are LeNet \cite{lecun1998gradient}, AlexNet \cite{krizhevsky2012imagenet}, VGG \cite{simonyan2014very}, GoogLeNet \cite{szegedy2015going}, CaffeNet \cite{jia2014caffe}, and ResNet \cite{he2016deep}. 
\end{itemize}

\subsection{Threat Models for Adversarial ML Attacks on CAVs}

Threat modeling is the procedure of answering a few common and straight forward questions related to the system being developed or deployed from a hypothetical attacker's point of view. Threat modeling is a fundamental component of security analysis. It requires that some fundamental questions related to the threat are addressed \cite{kevin}. In particular, a threat model should identify:

\begin{itemize}

\item the \textit{system principals}: what is the system and who are the stakeholders? 

\item  the \textit{system goals}: what does the system intend to do?

\item the \textit{system adversities}: what potential bad things can happen due to adverse situations or motivated adversaries?

\item  the \textit{system invariants}: what must be always true about the system even if bad things happen? 

\end{itemize}

The key goal of threat modeling is to optimize the security of the system by determining security goals, identifying potential threats, and vulnerabilities, and to develop countermeasures for preventing or mitigating their associated effects on the system. Answering these questions requires careful logical thoughts and significant expertise and time. 

As the focus of this paper is on highlighting the potential vulnerabilities of using ML techniques in CAVs, the scope of our study is restricted to the adversarial ML threat in CAVs. In the remainder of this section, we discuss the various facets of the adversarial ML threat in CAVs (a taxonomy aggregating these issues is illustrated in Figure \ref{fig:th_models}). 

%A fundamental challenge in security research for CAVs is to identify to what extent CAVs are capable of securely performing operations in adversarial conditions. Threat modeling is useful for systematically addressing this question. 

\subsubsection{Adversarial Attack Type}

% \jq{This information is not depicted in the figure. Try to make the text in this section and the information in the figure closely aligned.}

In the literature, the attacks on learning systems are generally categorized into three dimensions \cite{barreno2010security}: 

\begin{itemize}
    \item[(a)] \textit{Influence}: It includes causative (trying to get control over training data) and exploratory (exploiting mis-classifications of the model without affecting the training process) attacks.
    \item[(b)] \textit{Specificity}: It involves targeted and indiscriminate attacks on a specific instance.
    \item[(c)] \textit{Security Violation}: It is concerned with the integrity of assets and availability of the service attack. 
\end{itemize}

% \jq{See if this needs more description.}
The first dimension describes the capabilities of the adversary and whether the attacker has the ability to affect the learning by poisoning training data. Instead, the attacker exploits the model by sending new samples and observing their responses to get the intended behavior. The second axis indicates the specific intentions of the attacker, i.e., whether the attacker is interested in realizing a targeted attack on one particular sample or he aims to cause learned model t fail in an indiscriminate fashion. The third dimension detail the types of security violation an attacker can cause, e.g., the attacker may aim to bypass harmful messages to bypass through the filter as false negatives or realizing denial of service by causing benign samples misclassified as false positives.

%\subsubsection{Autonomous Vehicles: The ML Attack Surface}
%The attack surface for CAVs can be defined with respect to the ML pipeline for CAVs, which was introduced in Figure \ref{}. An adversary can try to manipulate the data collection and processing system, tamper the model, or its outputs. 

\subsubsection{Adversarial Knowledge}

Based on the adversarial knowledge available to the adversaries, the adversarial ML attacks are divided into three types; namely, \textit{white-box, gray-box}, and \textit{black-box} attacks. \textit{White-box} attacks assume complete knowledge about the underlying ML model including information about the optimization technique, the trained ML model, model architecture, activation function, hyper-parameters, layer weights, and training data.  Gray-box attacks assume a partial knowledge about the targeted model whereas the black-box adversarial attack assumes the adversary has zero knowledge and no access to the underlying ML model and the training data. Black box attack refers to the real-world knowledge where there is not much information about the targeted ML/DL scheme is available. In such cases, the adversary acts as a normal user and tries to infer from the output of the ML system. Black-box adversarial attacks make use of transferability property of adversarial examples where it is assumed that adversarial examples created for one ML/DL model will affect other models trained on datasets with a similar distribution to that of the original model \cite{tramer2017space}.

\subsubsection{Adversarial Capabilities}
Adversarial capabilities are important to be identified in security practice. As they define the strength of the adversaries to compromise the security of the system. In general, an adversary can be stronger or weaker based on the knowledge and access to the system. Adversarial capabilities advocate how and what type of attacks an adversary can realize using what type of attack vector on which attack surface. The attacks can be launched at two main phases; namely, inference and training. Inference time attacks are exploratory attacks that do not modify the underlying model. Instead they influence it to produce incorrect outputs. Inference attacks vary with the availability of system knowledge. The training time attacks aim at tampering with the model itself or influence its learning process and involve two types of attack methods \cite{papernot2016towards}. In the first type, adversarial examples are injected in the training data and in the second type, training data is directly modified.      

\subsubsection{Adversarial Specificity}
Another classification of the adversarial attacks is based on the specificity of the adversarial examples, where adversarial attacks are classified as \textit{targeted} and \textit{non-targeted} attacks. The attacks where adversarial perturbations are added to compromise the performance of a specific class in the data are known as the targeted adversarial attacks. Targeted adversarial attacks are launched by adversaries to create targeted misclassification (i.e., a specific road sign will be misclassified by the self-driving vehicle while the rest of the road sign classification system will function correctly) or source/target misclassification (i.e., a certain road traffic sign will be always classified in a pre-determined wrong class by the road sign classifier in a self-driving vehicle). Whereas adversarial perturbations created for deteriorating the performance of the model irrespective of any class of data are known as non-targeted adversarial attacks. Non-targeted attacks are launched by adversaries to reduce the classification confidence (i.e., a traffic sign will be detected with less accuracy which was previously detected with high accuracy) and misclassification (i.e., a road traffic sign will be classified in any other class than its original one). 

\subsubsection{Adversarial Falsification}
The adversary can attempt two types of falsification attacks; namely, false positive attacks, and false negative attacks \cite{yuan2019adversarial}. In the first attack, an adversary generates a negative sample which can be misclassified as a positive one. Let's assume such attack has been launched on the image classification system of an autonomous vehicle. A false positive will be an adversarial image predicted to be of a class with high confidence to whom it did not belong and is unrecognizable to humans. On the contrary, while attempting false negative attacks, the adversary generates a positive sample which can be misclassified as a negative one. In adversarial ML, this type of attack is referred to as an evasion attack.  

\subsubsection{Attack Frequency}
The adversarial attacks can be single step or consist of an iterative optimization process. Compared to single step attacks, iterative adversarial attacks are stronger; however, they require frequent interactions for querying the ML system and subsequently require a large amount of time and computational resources for their efficient generation.   

\subsubsection{Adversarial Goals}
The last component of the threat modeling is the articulation of the adversary's goals. The classical approach to model adversarial goals includes modeling of the adversary's desires to impact the confidentiality, integrity, and availability (known as the CIA model) and a fourth, yet important dimension is the privacy \cite{papernot2016towards}.    

% \textcolor{red}{The final component of the threat modeling is the articulation of the adversaries' goals. Traditionally, adversarial goals impact the confidentiality, integrity, and availability (CIA) model and a fourth, yet important dimension, is the model's privacy. }

%\section{Adversarial ML and CAVs}
%\label{sec:adv_ml}
%In this section, a detailed discussion on adversarial ML from the perspective of CAVs is provided.
%\subsection{Adversarial ML}

\begin{table*}[]
\centering
\scriptsize
\caption{Summary of the state-of-the-art attacks}
\scalebox{0.95}{
\begin{tabular}{|c|c|c|c|c|c|c|c|}
\hline
\textbf{Year} & \textbf{Authors} & \textbf{Method} & \textbf{Adversarial Knowledge} & \textbf{Adversarial Specificity} & \textbf{Perturbation Scope} & \textbf{Perturbation Norm} & \textbf{Attack Learning} \\ \hline
2014 & Szegedy et al. \cite{szegedy2013intriguing} & L-BFGS & White box & Targeted & Image specific & $L_{\infty}$ & One Shot \\ \hline
2015 & Goodfellow et al. \cite{goodfellow2014explaining} & FGSM & White box & Targeted & Image specific & $L_{\infty}$ & One shot \\ \hline
2016& Kurakin et al. \cite{kurakin2016adversarial} & BIM \& ILCM & White box & Non targeted & Image specific & $L_{\infty}$ & Iterative  \\ \hline
2016 & Papernot et al. \cite{papernot2016limitations} & JSMA & White box & Targeted & Image specific & $L_{0}$ & Iterative \\ \hline
2016 & Moosavi et al. \cite{moosavi2016deepfool} & DeepFool & White box & Non targeted & Image specific & $L_{2}$,$L_{\infty}$ & Iterative  \\ \hline
2017 & Carlini et al. \cite{carlini2017towards} & C\&W attacks & White box & Targeted & Image specific & $L_{0}$,$L_{2}$,$L_{\infty}$ & Iterative  \\ \hline
2017 & Moosavi et al. \cite{moosavi2017universal} & Uni. perturbations & White box & Non targeted & Universal & $L_{2}$,$L_{\infty}$ & Iterative  \\ \hline
2017 & Sarkar et al. \cite{sarkar2017upset} & UPSET & Black box & Targeted & Universal & $L_{\infty}$ & Iterative \\ \hline
2017 & Sarkar et al. \cite{sarkar2017upset} &  ANGRI & Black box & Targeted & Image specific & $L_{\infty}$ & Iterative \\ \hline

2017 & Cisse et al. \cite{cisse2017houdini} & Houdini & Black box & Targeted & Image specific & $L_{2}$,$L_{\infty}$ & Iterative \\ \hline
2018 & Baluja et al. \cite{baluja2018learning} & ATNs & White box & Targeted & Image specific & $L_{\infty}$ & Iterative \\ \hline
2019 & Su et al. \cite{su2019one} & One-pixel & Black box & Non Targeted & Image specific & $L_{0}$ & Iterative  \\ \hline
\end{tabular}}
\label{tab:summary_attack_methods}
\end{table*}

\subsection{Review of Existing Adversarial ML Attacks}

% \jq{Have a hierarchy in this subsection and use subsubsections}.
\subsubsection{Adversarial ML Attacks on Conventional ML Schemes}
A pioneering work on adversarial ML was performed by Dalvi et al. \cite{dalvi2004adversarial} in 2004 where they proposed a minimum distance evasion of the linear classifier and tested there proposed attack on spam classification system highlighting the threat of adversarial ML examples. A similar contribution was made by Lowd et al. \cite{lowd2005adversarial} in 2005 where they proposed adversarial classifier reverse engineering technique for constructing an adversarial attack on classification problems. In 2006, Barreno et al. \cite{barreno2006can} discussed the security of ML in adversarial environments and provided a taxonomy of attacks on ML schemes along with the potential defenses against them. In 2010, Huang et al. \cite{huang2011adversarial} provided the first consolidated review of adversarial ML where they discussed the limitations on the classifiers and adversaries in real-world settings. Biggio et al. \cite{biggio2012poisoning} proposed poisoning attack on Support Vector Machines (SVM) to increase the test error in SVM, their attack successfully altered the test error of SVM with linear and non-linear kernels. The same authors also proposed an evasion attack where they used a gradient-based approach for evading PDF malware detectors \cite{biggio2013evasion} and tested their  attack on SVM and simple neural networks.

\begin{table*}[]
\centering
\caption{Accidents caused by self-driving vehicles due to unintended adversarial conditions}
\scalebox{0.9}{
\begin{tabular}{|c|c|l|l|l|}
\hline
\textbf{Year} & \textbf{Company} & \multicolumn{1}{c|}{\textbf{Cause of the accident}} & \multicolumn{1}{c|}{\textbf{Damages}} & \multicolumn{1}{c|}{\textbf{System failure}} \\ \hline
2014 & Hyundai & Weather (Rain fall) & Car crashed & Camera object detection failure \\ \hline
2016 & Google Waymo & Speed estimation failure & Car crashed in the bus & Dynamic object movement detection failure \\ \hline
2016 & Tesla & Image classification and image contrast failure & \begin{tabular}[c]{@{}l@{}}Car crashed in the neighboring truck\\ and the driver was killed\end{tabular} & \begin{tabular}[c]{@{}l@{}}Camera's detection and \\ classification suite failure\end{tabular} \\ \hline
2017 & Uber & \begin{tabular}[c]{@{}l@{}}Overreaction to an unseen event\\ (Near by accident)\end{tabular} & Car crashed & Lack of robustness in control system \\ \hline
2017 & General Motors & \begin{tabular}[c]{@{}l@{}}Stuck in a dilemma \\ (Lane change decision reversal)\end{tabular} & Car knocked over a motorcyclist & Coordination and state estimation failure \\ \hline
2018 & Uber & \begin{tabular}[c]{@{}l@{}}Confusion in the software decision system\\  and safety system failure\end{tabular} & Killed a person on the road & Failure of planning and perception system \\ \hline
\end{tabular}
}
\label{acc}
\end{table*}

\subsubsection{Adversarial ML Attacks on DNNs}
Adversarial ML attacks on DNNs were first observed by Szegedy et al. \cite{szegedy2013intriguing} where they demonstrated that DNNs can be fooled by minimally perturbing their input images at test time, the proposed attack was a gradient-based attack where minimum distance based adversarial examples were crafted to fool the image classifiers. Another gradient-based attack was proposed by Goodfellow et al. \cite{goodfellow2014explaining}. In this attack, they formulated adversarial ML as a min-max problem and adversarial examples were produced by calculating the lower bound on the adversarial perturbations. This method was termed as FGSM and is still considered a very effective algorithm for creating adversarial examples. Adversarial training was also introduced in the same paper as a defensive mechanism against adversarial examples. 
Kurakin et al. \cite{kurakin2016adversarial} highlighted the fragility of ML/DL schemes in real-world settings using images taken from a cell phone camera for adversarial example generation. The adversarial samples were created by using the basic iterative method (BIM) an extended version of FGSM. The resultant adversarial examples were able to fool the state-of-art image classifier. In \cite{engstrom2017rotation}, authors demonstrated that only rotation and translation are sufficient for fooling state-of-the-art deep learning based image classification models, i.e., convolutional neural networks(CNNs). In a similar study \cite{pei2017towards}, ten state-of-the-art DNNs were shown to be fragile to the basic geometric transformation, e.g., translation, rotation, and blurring. Liu et al. presented a trojaning attack on neural networks that works by modifying the neurons of the trained model instead of affecting the training process \cite{LiuMALZW018}. Authors used trojan as a backdoor to control the trojaned ML model as desired and tested it on an autonomous vehicle. The car misbehaves when a specific billboard (trojan trigger) is encountered by it on the roadside.

Papernot et al. \cite{papernot2016limitations} exploited the mapping between the input and output of  DNNs to construct a white-box Jacobian saliency-based adversarial attack (JSMA) scheme to fool the DNN classifiers. The same authors also proposed another defense against adversarial perturbations by using defensive distillation. Defensive distillation is a training method in which a model is trained to predict the classification probabilities of another model which was trained on the baseline standard to give more importance to accuracy. Papernot et al. \cite{papernot2017practical} also proposed a black-box adversarial ML attack where they exploited the transferability property of adversarial examples to fool the ML/DL classifiers. This black-box adversarial attack was based on the substitute model training which not only fools the ML/DL classifiers but also breaks the distillation defensive mechanism. Carlini et al. \cite{carlini2017towards} proposed a suite of three adversarial attacks termed as C\&W attacks on DNNs by exploiting three distinct distance measures $L{_1}, L{_2}$, and $L{_\infty}$. These attacks have not only evaded the DNN classifiers but also evaded the defensive distillation successfully. This demonstrated that defensive distillation is not an appropriate method for building robustness. In another paper, Carlini et al. \cite{carlini2017adversarial} presented that the proposed adversarial attacks in \cite{carlini2017towards} have successfully evaded the ten well known defensive schemes against adversarial examples. Right now these attacks are also considered as state-of-art adversarial ML attacks. Furthermore, Carlini et al. successfully demonstrated an adversarial attack on speech recognition system by adding small noise in the audio signal that forces the underlying ML model to generate intended commands/text \cite{carlini2018audio}. In \cite{brown2017adversarial}, an adversarial patch affixed to an original image forces the deep model to mis-classify that image. Such universal targeted patches fool classifiers without requiring knowledge of the other items in the scene. Sich patches can be created offline and then broadly shared. More details on adversarial ML attacks can be found in \cite{yuan2019adversarial, wang2019survey, chakraborty2018adversarial, vorobeychik2018adversarial, biggio2018wild, akhtar2018threat}. A summary of different state-of-the-art adversarial perturbation generation methods is provided in Table \ref{tab:summary_attack_methods}.

\begin{table}[!ht]
\centering
\caption{Domains affected by adversarial machine learning (ML) and their applications}
\begin{tabular}{|c|l|l|}
\hline
\textbf{Domain} & \textbf{Application} & \textbf{Papers} \\ \hline
\multirow{6}{*}{Imaging} & Digit Recognition & \cite{papernot2016limitations}, \cite{goodfellow2014explaining}, \cite{carlini2017towards}, ...  \\ \cline{2-3} 
 & Object Detection & \cite{carlini2017towards}, \cite{papernot2017practical}, \cite{balda2018generation}, ... \\ \cline{2-3} 
 & Traffic Signs Recognition & \cite{papernot2017practical}, \cite{sitawarin2018darts}, \cite{sitawarin2018rogue}, ... \\ \cline{2-3} 
 & Semantic Segmentation & \cite{arnab2018robustness}, \cite{metzen2017universal}, \cite{fischer2017adversarial}, ...\\ \cline{2-3} 
 & Reinforcement Learning & \cite{tretschk2018sequential}, \cite{pattanaik2018robust}, \cite{lin2017tactics}, ... \\ \cline{2-3} 
 & Generative Modeling & \cite{kos2018adversarial}, \cite{gondim2018adversarial}, \cite{pasquini2019out}, ... \\ \hline
\multirow{3}{*}{Text} & Text Classification & \cite{sato2018interpretable}, \cite{miyato2016adversarial}, \cite{li2018textbugger}, ... \\ \cline{2-3} 
 & Sentiment Analysis & \cite{papernot2016crafting}, \cite{li2018textbugger}  \\ \cline{2-3} 
 & Reading Comprehension & \cite{jia2017adversarial}, \cite{mudrakarta2018did} \\ \hline
\multirow{4}{*}{Networking} & Intrusion Detection & \cite{corona2013adversarial}, \cite{lin2018idsgan}, \cite{wang2018deep}, ...  \\ \cline{2-3} 
 & Anomaly Detection & \cite{salem2018anomaly}, \cite{wang2014man} \\ \cline{2-3} 
 & Malware Classification & \cite{grosse2016adversarial}, \cite{kolosnjaji2018adversarial}, \cite{usama2018adversarial}, ...  \\ \cline{2-3} 
 & Traffic Classification & \cite{ahmed2017poster}, \cite{viegas2017stream}  \\ \hline
Audio & Speech Recognition & \cite{carlini2018audio}, \cite{schonherr2018adversarial}, \cite{cisse2017houdini}, ... \\ \hline
\end{tabular}
\label{tab:apps}
\end{table}

\subsection{Adversarial ML attacks on CAVs}

% \jq{Have a hierarchy in this subsection and use subsubsections}.

ML and DL act are core ingredients for performing many key tasks in self-driving vehicles. Beyond providing deeply embedded information for the decision making process within the vehicle's components, they also play an important role in V2I and V2V, and V2X communications. As described in earlier sections, ML/DL schemes are very vulnerable to small carefully crafted adversarial perturbations. Self-driving vehicles are also threatened by this security risk along with other traditional security risks. Adversarial ML has affected many application domains including imaging, text, networking, and audio as highlighted in Table \ref{tab:apps}. 

\subsubsection{Autonomous Vehicles Accidents Due to Unintended Adversarial Conditions}
The autonomous vehicles developed so far are not robust to unintended adversarial conditions and there have been few reported fatalities caused by the malfunctioning of DNN-based autonomous vehicles where adversarial examples were unintentionally created by the DNN operating the autonomous vehicle. In 2014, during Hyundai competition, an autonomous vehicle crashed because of a sensor failure due to shifting in the angle of the car and direction of the sun\footnote{\url{https://bit.ly/2SWLxUY}}. Another incident was reported in 2016 where a Tesla autopilot was not able to handle the image contrast which resulted in the death of the driver\footnote{\url{https://cnnmon.ie/2VOB283}}. It was also reported that the Tesla autopilot unable to differentiate between the bright sky and a white truck which resulted in a horrible accident. A similar accident happened to Google self-driving car where the car was unable to estimate the relative speed which resulted in a collision with a bus\footnote{\url{https://bit.ly/1U0O6yx}}. In 2018 Uber self-driving car also faced an accident due to malfunctioning in the DNN-based system which resulted in a pedestrian fatality\footnote{\url{https://bit.ly/2SWmb9N}}. Table \ref{acc} provides a detailed description of accidents caused by malfunctions in different components of self-driving vehicles.

%and Table \ref{acc} presents a summary of accidents caused by self-driving cars due to unintended conditions.   

\subsubsection{Physical World Attacks on Autonomous Vehicles}
Aung et al. \cite{aung2017building} used FGSM and JSMA schemes to generate adversarial traffic signs to successfully evade the DNN-based traffic sign detection schemes to highlight the problem of adversarial examples in autonomous driving. Sitawarin et al. \cite{sitawarin2018rogue} proposed a real-world adversarial ML attack by altering the traffic signs and logos with adversarial perturbations while keeping the visual perception of the traffic and logo signs. In another work, Sitawarin et al. \cite{sitawarin2018darts} proposed a technique for generating out-of-distribution adversarial examples to perform an evasion attack on ML-based sign recognition system of autonomous vehicles. They also proposed a Lenticular printing attack where they exploited the camera height in autonomous vehicles to create an illusion of false traffic signs in the physical environment to fool the sign recognition system of autonomous vehicles.

Object detection is another integral part of the perception module of autonomous vehicles where state-of-the-art DNN-based schemes such as Mask R-CNN \cite{he2017mask} and YOLO \cite{redmon2018yolov3} are used for object detection. Zhang et al. \cite{zhang2018camou} proposed a camouflage physical world adversarial attack by approximately imitating how a simulator applies camouflage to the vehicle and then minimized the approximated detection score by using local search for optimal camouflage. The proposed adversarial attack successfully fooled image-based object detection systems. Another physical world adversarial example generation scheme on object detection is performed by Song et al. \cite{song2018physical} where the perturbed ``STOP'' sign remained hidden from the state-of-art object detectors like Mask R-CNN and YOLO. They produced adversarial perturbations by the robust physical perturbations (RP$_{2}$) \cite{eykholt2018robust} algorithm. Recently Zhou et al. \cite{zhou2018deepbillboard} proposed DeepBillboard a systematic way for generating adversarial advertisement billboards to inject a malfunction in the steering angle of the autonomous vehicle. The proposed adversarial billboard misled the average steering angle by 26.44 degrees. Table \ref{table3} provides a summary of state-of-the-art adversarial attacks on self-driving vehicles. In a recent study \cite{bansal2018chauffeurnet}, imitation learning has been shown robust enough for autonomous vehicles to drive in a realistic environment. Authors proposed a model named ChauffeurNet that learns to drive the vehicle by imitating best and synthesizing worst.

\begin{table*}[]
\centering
\caption{Adversarial attacks on self-driving vehicles: summary of state-of-the-art}
%\tiny
\scalebox{0.74}{
\begin{tabular}{|c|l|l|l|l|l|}
\hline
\multicolumn{1}{|l|}{\textbf{Attack Objective}} & \textbf{Specific work} & \textbf{Problem formulation} & \textbf{Data} & \textbf{Threat model} & \textbf{Attack results} \\ \hline
\begin{tabular}[c]{@{}c@{}}Perception system\\  failure\end{tabular} & \begin{tabular}[c]{@{}l@{}}DARTS \cite{sitawarin2018darts}: Traffic signs\\ manipulation\end{tabular} & \begin{tabular}[c]{@{}l@{}}Generating adversarial examples for\\ CNN-based traffic sign detection by\\ performing out-of-distribution attacks\\ along with Lenticular printing.\end{tabular} & \begin{tabular}[c]{@{}l@{}}GTSRB \\ GTSDB\end{tabular} & \begin{tabular}[c]{@{}l@{}}1) Virtual and physical world attack\\ 2) White and black-box mode\end{tabular} & \begin{tabular}[c]{@{}l@{}}DARTS has successfully \\ fooled the perception system\\ of self-driving car.\end{tabular} \\ \hline
\begin{tabular}[c]{@{}c@{}}Perception system\\  failure\end{tabular} & \begin{tabular}[c]{@{}l@{}}Rogue Signs \cite{sitawarin2018rogue}: Traffic signs \\ and logos manipulation\end{tabular} & \begin{tabular}[c]{@{}l@{}}End-to-end pipeline for adversarial \\ example generation for CNN-based \\ traffic signs and logo detection.\end{tabular} & GTSRB & \begin{tabular}[c]{@{}l@{}}1) Virtual and physical world attack\\ 2) White-box mode\end{tabular} & \begin{tabular}[c]{@{}l@{}}Fooled the perception system\\ of self-driving car with a success\\ rate of 99.7\%.\end{tabular} \\ \hline
\begin{tabular}[c]{@{}c@{}}Object detection\\ failure\end{tabular} & \begin{tabular}[c]{@{}l@{}}ShapeShifter \cite{chen2018shapeshifter}: Adversarial\\ attack on Faster R-CNN\end{tabular} & \begin{tabular}[c]{@{}l@{}}Adversarial attack on bounding boxes\\ of Faster R-CNN by using expectation \\ over transformation techniques.\end{tabular} & MS-COCO & \begin{tabular}[c]{@{}l@{}}1) Virtual and physical world attack\\ 2) White-box mode\end{tabular} & \begin{tabular}[c]{@{}l@{}}Caused malfunction in Faster\\ R-CNN of self-driving car\\ with 93\% success.\end{tabular} \\ \hline
\begin{tabular}[c]{@{}c@{}}Motion planning and \\ perception system \\ failure\end{tabular} & \begin{tabular}[c]{@{}l@{}}DeepBillboard \cite{zhou2018deepbillboard}: Adversarial \\ attack through drive-by\\ billboards.\end{tabular} & \begin{tabular}[c]{@{}l@{}}Adversarial attack on steering angle \\ of the self-driving car by using adversarial \\ perturbations in drive-by billboards.\end{tabular} & \begin{tabular}[c]{@{}l@{}}1) Udacity self-driving\\  car challenge dataset\\ 2) Dave testing dataset\\ 3) Kitti dataset\end{tabular} & \begin{tabular}[c]{@{}l@{}}1) Virtual and physical world attack\\ 2) White and black-box mode\end{tabular} & \begin{tabular}[c]{@{}l@{}}Caused a 23 degree malfunction\\ in steering angle of \\ self-driving car.\end{tabular} \\ \hline
\begin{tabular}[c]{@{}c@{}}Object detection\\  failure\end{tabular} & \begin{tabular}[c]{@{}l@{}}CAMOU \cite{zhang2018camou}: Adversarial\\ camouflage to fool the\\ object detector\end{tabular} & \begin{tabular}[c]{@{}l@{}}Generating adversarial perturbation \\ to prevent the self-driving car from\\ Mask R-CNN-based object detector.\end{tabular} & \begin{tabular}[c]{@{}l@{}}Unreal engine \\ simulator\end{tabular} & \begin{tabular}[c]{@{}l@{}}1) Virtual and physical world attack\\ 2) White and black-box mode\end{tabular} & \begin{tabular}[c]{@{}l@{}}Caused a 32.74\% drop\\ in the performance of\\ Mask R-CNN.\end{tabular} \\ \hline
\begin{tabular}[c]{@{}c@{}}Object detection\\  failure\end{tabular} & \begin{tabular}[c]{@{}l@{}}Song et al. \cite{song2018physical}: Object \\ disappearance and\\ creation attack\end{tabular} & \begin{tabular}[c]{@{}l@{}}Generating adversarial perturbation\\ based stickers where object detection\\ schemes like Yolo and R-CNN used\\ in self-driving cars fails to recognize\\ certain signs and logos. Furthermore\\ object detection schemes start\\ detecting things that are not present \\ in the frame.\end{tabular} & Video of traffic signs & \begin{tabular}[c]{@{}l@{}}1) Virtual and physical world attack\\ 2) White-box mode\end{tabular} & \begin{tabular}[c]{@{}l@{}}The object detection schemes \\ fails to recognize traffic signs \\ nearly in 86\% of the frames in \\ the video.\end{tabular} \\ \hline
\begin{tabular}[c]{@{}c@{}}Perception system\\ failure\end{tabular} & \begin{tabular}[c]{@{}l@{}}Eykholt et al. \cite{eykholt2018robust}: Robust \\ physical perturbations\\ against visual classification\\ under different physical\\ environment.\end{tabular} & \begin{tabular}[c]{@{}l@{}}Robust physical perturbations are \\ created to fool LISA-CNN and \\ GTSRB-CNN based traffic sign\\ classification schemes.\end{tabular} & \begin{tabular}[c]{@{}l@{}}1) LISA \\ 2) GTSRB\end{tabular} & \begin{tabular}[c]{@{}l@{}}1) Virtual and physical world attack\\ 2) White and black-box mode\end{tabular} & \begin{tabular}[c]{@{}l@{}}In few cases caused 100\%\\ performance drop in visual \\ classification of traffic signs\end{tabular} \\ \hline
\begin{tabular}[c]{@{}c@{}}Perception and \\ controller system\\ failure\end{tabular} & \begin{tabular}[c]{@{}l@{}}Tuncali et al. \cite{ tuncali2018simulation}: Simulation\\ based adversarial test \\ generation for self-driving\\ cars.\end{tabular} & \begin{tabular}[c]{@{}l@{}}Testing and verifying self-driving\\ car's perception and controller\\ system against adversarial examples.\end{tabular} & \begin{tabular}[c]{@{}l@{}}Simulated data\\ from proposed \\ simulator\end{tabular} & \begin{tabular}[c]{@{}l@{}}1) Virtual environment \\ 2) White and black-box mode\end{tabular} & \begin{tabular}[c]{@{}l@{}}Designed system was able \\ to detect critical cases in \\ autonomous car's perception\\ and control.\end{tabular} \\ \hline
\begin{tabular}[c]{@{}c@{}}Controller system\\ failure\end{tabular} & \begin{tabular}[c]{@{}l@{}}Yaghoubi et al. \cite{ yaghoubi2018gray}: Finding\\ gray-box adversarial examples\\ for closed loop autonomous \\ cars control system.\end{tabular} & \begin{tabular}[c]{@{}l@{}}Testing the controller and perception\\ system of self-driving cars against\\ gradient-based gray-box adversarial\\ examples.\end{tabular} & Simulated data & \begin{tabular}[c]{@{}l@{}}1) Virtual environment \\ 2) Gray box mode\end{tabular} & \begin{tabular}[c]{@{}l@{}}Gray-box adversarial examples \\ have outperformed simulated\\ Annealing optimization in \\ a dummy control system\\ problem.\end{tabular} \\ \hline
\begin{tabular}[c]{@{}c@{}}End-to-end \\ autonomous control\\  failure\end{tabular} & \begin{tabular}[c]{@{}l@{}}Boloor et al. \cite{ boloor2019simple }: Physical\\ adversarial examples \\ against E2E driving\\ models.\end{tabular} & \begin{tabular}[c]{@{}l@{}}Disrupting steering by using physical\\ perturbation in the environment.\end{tabular} & \begin{tabular}[c]{@{}l@{}}CARLA simulated\\ data\end{tabular} & \begin{tabular}[c]{@{}l@{}}1) Virtual environment\\ 2) White-box mode\end{tabular} & \begin{tabular}[c]{@{}l@{}}Physical adversarial\\ perturbation has forced the\\ self-driving car to crash.\end{tabular} \\ \hline
\end{tabular}
}
\label{table3}
\end{table*}

\section{Towards Developing Adversarially Robust ML Solutions}
\label{sec:robust_sols}
%\jq{Highlight not only problems and open issues but also to identify the major solution types proposed in literature.}
As discussed above, despite the outstanding performance of ML techniques in many settings, including human level accuracy at recognizing images. These techniques exhibit strict vulnerability to carefully crafted adversarial examples. In this section, we present an outline of approaches for developing adversarially robust ML solutions. We define the robustness as the ability of the ML model to restrain adversarial examples.

In the literature, defenses against adversarial attacks have been divided into two broad categories: (1) \textit{reactive} detect adversarial observations (input) after deep models are trained; and (2) \textit{proactive} make the deep model robust against adversarial examples before the attack. 

\begin{figure*}[]
    \centering
    \includegraphics[width=0.98\textwidth]{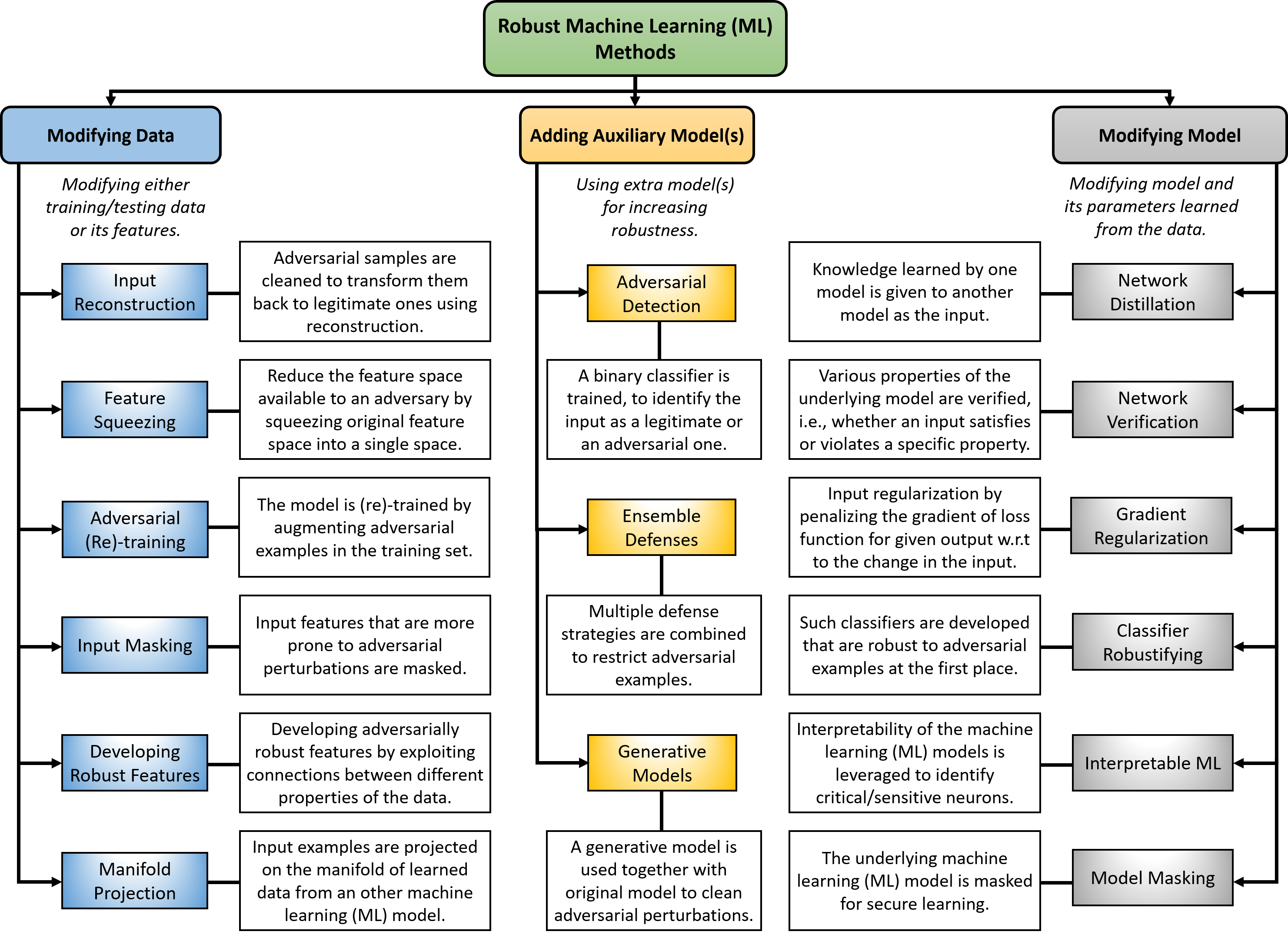}
    \caption{Taxonomy of robust machine learning (ML) methods categorized into three classes: (1) Modifying Data (2) Adding Auxiliary Model(s) and (3) Modifying Models.} 
    \label{fig:taxonomy}
\end{figure*}

Alternatively, these techniques can also be broadly divided into three categories: (1) modifying data; (2) adding auxiliary models; and (3) modifying models. The reader is referred to Figure \ref{fig:taxonomy} for a visual depiction of a taxonomy of robust ML solutions in which various techniques that fall in these categories are also listed. These categories are detailed next. 

  %Following are few approaches to make ML models robust against adversarial attacks.   

\subsection{Modifying Data}
The methods falling under this category mainly deal with modification of either the training data (e.g.,
adversarial retraining) and its features or test data (e.g., data pre-processing). Widely used approaches that utilize such methods are described below.  

\subsubsection{Adversarial (Re-)training}
The training with adversarial examples has been firstly proposed by Goodfellow et al. \cite{goodfellow2014explaining} and Huang et al. \cite{huang2015learning} as a defense strategy to make deep neural networks (DNNs) robust against adversarial attacks. They trained the model by augmenting adversarial examples in the training set. Furthermore, Goodfellow et al. showed that adversarial training could provide better regularization for DNNs. In \cite{goodfellow2014explaining,huang2015learning}, the adversarial robustness of ML models was evaluated on the MNIST dataset having 10 classes while in \cite{katz2017reluplex}, a comprehensive evaluation of adversarial training was performed on a considerably large dataset, i.e., ImageNet having 1000 classes. The authors used 50\% of the dataset for adversarial training and this strategy increased the robustness of DNNs for single step adversarial attack (e.g., FGSM \cite{goodfellow2014explaining}). However, the strategy failed for iterative adversarial examples generation methods such as the basic iterative method (BIM) \cite{kurakin2016adversarial}.

\subsubsection{Input Reconstruction}
The idea of input reconstruction is to clean the adversarial examples to transform them back to legitimate ones. Once the adversarial examples have been transformed, they will not affect the prediction of DNN models. For robustifying DNN, a technique named deep contractive autoencoder has been proposed in \cite{gu2014towards}. They trained a denoising autoencoder for cleaning adversarial perturbations. 

%In a similar study \cite{song2017pixeldefend}, authors devised a CNN based technique for reconstructing adversarial images back to the original one. They named their defense system PixelDefend that works by modifying all pixels values in each channel for maximizing the probability distribution of training set.  

\subsubsection{Feature Squeezing}
Xu et al. \cite{xu2017feature} leveraged the observation that input feature spaces are typically unnecessarily large and provide a vast room for an adversary to construct adversarial perturbations and thereby proposed feature squeezing as a defense strategy to adversarial examples. The available feature space to an adversary can be reduced using feature squeezing that combines samples having heterogeneous feature vectors in the original space into a single space. They perform feature squeezing at two levels: (1) reducing color bit depth; (2) spatial domain smoothing using both local and non-local method. Also, they evaluated eleven state-of-the-art adversarial perturbations generation methods on three different datasets, i.e., MNIST, CIFAR-10, and ImageNet. However, this defense strategy was found to be less effective in a later study \cite{he2017adversarial}. 

\subsubsection{Features Masking}
In \cite{gao2017deepcloak}, authors proposed to add a masking layer before the softmax layer of the classifier that is mainly responsible for the classification task. The purpose of adding the masking layer was to mask the most sensitive features of input that are more prone to adversarial perturbations by forcing the corresponding weights of this layer to zero.

\subsubsection{Developing Adversarially Robust Features}
This method has been recently proposed as an effective approach to make DNNs resilient against adversarial attacks \cite{garg2018spectral}. Authors leveraged the connections between the natural spectral geometrical property of the dataset and the metric of interest for developing adversarially robust features. They empirically demonstrated that the spectral approach can be effectively used to generate adversarially robust features that can be ultimately used to develop robust models.   

\subsubsection{Manifold Projection}
In this method, input examples are projected on the manifold of learned data from another ML model, generally, the manifold is provided by a generative model. For instance, Song et al. \cite{song2017pixeldefend} leveraged generative models to clean the adversarial perturbations from malicious images and then the cleaned images are given to the non-modified ML model. Furthermore, this paper ascertains that regardless of the attack type and targeted model, the adversarial examples lie in the low probability regions of the training data distribution. In a similar study \cite{shen2017ape}, authors used  generative adversarial networks (GANs) for cleaning adversarial perturbations. Similarly, Meng et al. proposed a framework named MagNet that includes one or more detectors and a reformer network \cite{meng2017magnet}. The detector network is used to classify normal and adversarial examples by learning the manifold of normal examples, whereas, the reformer network moves adversarial examples towards the learned manifold. 

\subsection{Modifying Model}
The methods that fall in this category mainly modify the parameters/features learned by the trained model (e.g., defensive distillation), a few prominent such methods are described next.  

\subsubsection{Network Distillation}
Papernot et al. \cite{papernot2016distillation} adopted network distillation as a procedure to defend against adversarial attacks. The notion of distillation was originally proposed by Hinton et al. \cite{hinton2015distilling} as a mechanism for effectively transferring knowledge from a larger network to a smaller one. The defense method developed by Papernot et al. uses the probability distribution vector generated by the first model as an input to the original DNN model. This increases the resilience of the DNN model towards very small perturbations. However, Carlini et al. showed that the defensive distillation method does not work against their proposed attack \cite{carlini2017adversarial}. 

\subsubsection{Network Verification}
Network verification aims to verify the properties of DNN, i.e., whether an input satisfies or violates certain property because it may restrain new unseen adversarial perturbations. For instance, a network verification method for robustifying DNN models using ReLU activation is presented in \cite{katz2017reluplex}. To verify the properties of the deep model, the authors used the satisfiability modulo theory (SMT) solver and showed that the network verification problem is NP-complete. The assumption of using ReLU with certain modifications is addressed in \cite{carlini2018ground}.

\subsubsection{Gradient Regularization}
Ross et al. \cite{ross2018improving} proposed using input gradient regularization as a defense strategy against adversarial attacks. In the proposed approach, they used differentiable DNN models and penalized the variation that results in the output with a change in the input. As a result, adversarial examples with small perturbations were unlikely to modify the output of deep models but this increases the training complexity with a factor of two. The notion of penalizing the gradient of loss function of models with respect to the inputs for robustification has been already been investigated in \cite{lyu2015unified}. 

\subsubsection{Classifier Robustifying}
In this method, classification models that are robust to adversarial attacks are designed from the ground up instead of detecting or transforming them. Bradshaw et al. \cite{bradshaw2017adversarial} utilized the uncertainty around the adversarial examples and developed a hybrid model using Gaussian processes (GPs) with RBF kernels on top of DNNs and showed that their approach is robust against adversarial attacks. The latent variable in GPs is expressed using a Gaussian distribution and is parameterized by mean and covariance and encoded with RBF kernels. Schott et al. \cite{schott2018towards} proposed the first adversarially robust classifier for MNIST dataset, where robustness is achieved by using analysis by synthesis through learned class-conditional data distribution. This work highlights the lack of research that provides guaranteed robustness against adversarial attacks.    

\subsubsection{Explainable and Interpretable ML}
In a recent study \cite{tao2018attacks}, an adversarial example detection approach is presented for a face recognition task that leverages the interpretability of DNN models. The key in this approach is the identification of critical neurons for an individual task that is performed by establishing a bi-directional correspondence inference between the neurons of a DNN model and its attributes. Then the activation values of these neurons are amplified to augment the reasoning part and the values of other neurons are decreased to conceal the uninterpretable part. Recently, Nicholas Carlini showed that this approach does not defend against untargeted adversarial perturbations generated using $L_\infty$ norm with a bound of 0.01 \cite{carlini2019ami}. 

\subsubsection{Masking ML Model}
In a recent study \cite{nguyen2018learning}, authors formulated the problem of adversarial ML as learning and masking problem and presented a classifier masking method for secure learning. To mask the deep model, they introduced noise in the DNN's logit output that was able to defend against low distortion attacks.

\subsection{Adding Auxiliary Model(s)}
These methods aim to utilize additional ML models to enhance the robustness of the main model (e.g., using
generative models for adversarial detection), such widely used methods are described as follows.

\subsubsection{Adversarial Detection}
In adversarial detection strategy, a binary classifier (detector) is trained, e.g., DNN to identify the input as a legitimate or an adversarial one \cite{lu2017safetynet,gopinath2017deepsafe}. In \cite{metzen2017detecting}, authors used a simple DNN-based binary adversarial detector as an auxiliary network to the main model. In a similar study \cite{katz2017towards}, authors introduced an outlier class while training the DNN model, the model then detects the adversarial examples by classifying them as an outlier. This defense approach has been used in a number of studies in the literature.  

\subsubsection{Ensembling Defenses}
As adversarial examples can be developed in a multi-facet fashion, therefore, multiple defense methods can be combined together (parallelly or sequentially) to defend against them \cite{kurakin2018ensemble}. PixelDefend \cite{song2017pixeldefend} is a prime example of ensemble defense in which an adversarial detector and an ``input reconstructor'' are integrated to restrain adversarial examples. However, He et al. showed that an ensemble of weak defense strategies does not provide a strong defense to adversarial attacks \cite{he2017adversarial}. Further, they demonstrated that adaptive adversarial examples transfer across several defense or detection proposals. 

\subsubsection{Using Generative ML Models}
Goodfellow et al. \cite{goodfellow2014explaining} firstly coined the idea of using generative training to defend adversarial attacks, however, in the same study they argued that being generative is not sufficient and presented an alternative hypothesis of ensemble training that works by ensembling multiple instances of original DNN models. In \cite{santhanam2018defending}, an approach named cowboy is presented to detect and defend against adversarial examples. They transformed adversarial samples back to data manifold by cleaning them using a GAN trained on the same dataset. Furthermore, authors empirically showed that adversarial examples lie outside the data manifold learned by the GAN, i.e., the discriminator of GAN consistently scores the adversarial perturbations lower than the real samples across multiple attacks and datasets. In another similar study \cite{samangouei2018defense}, a GAN-based framework named Defense-GAN is trained for modeling the distribution of legitimate images. During inference time, Defense-GAN finds a similar output without adversarial perturbations that is then fed to the original classifier. Also, the authors of both of these studies claimed that their method is independent of the DNN model and attack type and that it can be used in existing settings. The summary of various state-of-the-art adversarial defense studies is presented in Table \ref{tab:summary_defenses}. 

\begin{figure}[!ht]
    \centering
    \includegraphics[width=0.45\textwidth]{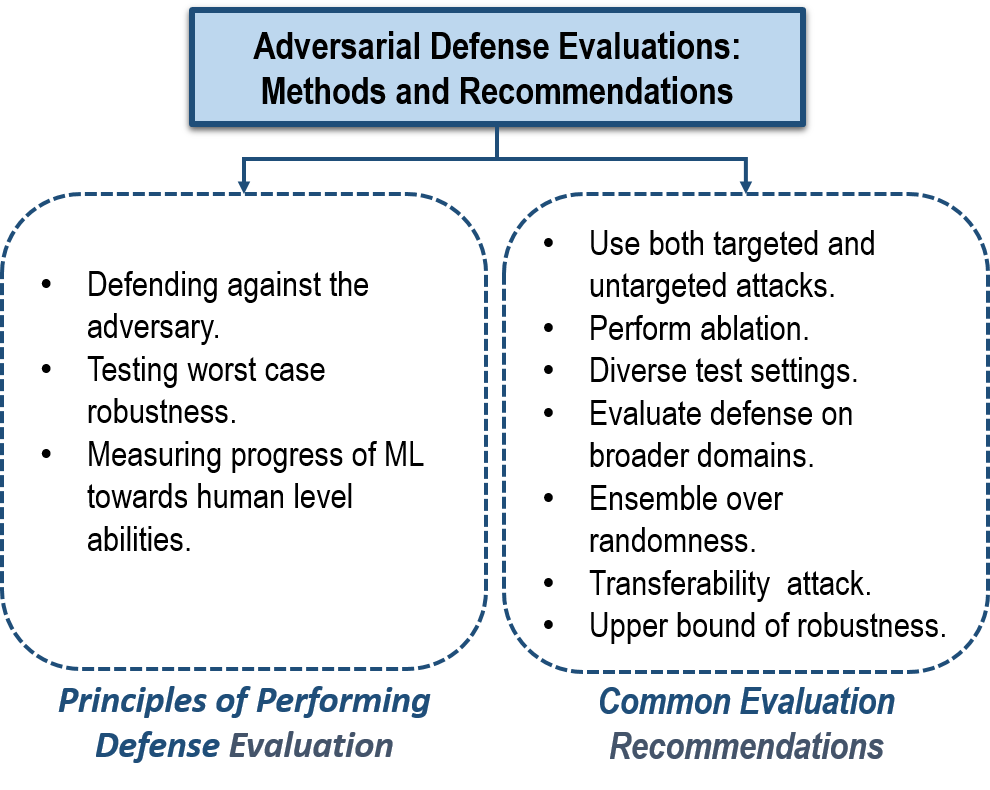}
    \caption{The taxonomy of different adversarial defense evaluation methods and recommendations.} 
    \label{fig:eval_methods}
\end{figure}

% \subsection{Miscellaneous Approaches}
% Wang et al. designed a statistical adversarial perturbations detection algorithm using mutation analysis \cite{wang2018detecting}.

\begin{table*}[]
\centering
\scriptsize
\caption{Summary of state-of-the-art adversarial defense approaches}
\scalebox{0.97}{
\begin{tabular}{|p{11mm}|p{11mm}|p{25mm}|p{25mm}|p{16mm}|p{12mm}|p{11mm}|p{13mm}|p{22mm}|}
\hline
\textbf{Reference} & \textbf{Type} & \textbf{Defense} & \textbf{Adv. Perturbation} & \textbf{Dataset} & \textbf{Threat Model} & \textbf{Original Accuracy} & \textbf{Adversarial Accuracy} & \textbf{Defense Success} \\ \hline
Gu et al. \cite{gu2014towards} & \multirow{38}{11mm}{Modifying Data} & Adversarial examples cleaning using denoising autoencoders (DAEs). & Local perturbations, e.g., additive Gaussian noise. & MNIST & Not clearly articulated & 99 (\%) & 100 (\%) & 99.1\% \\ \cline{1-1} \cline{3-9} 
Xu  et  al. \cite{xu2017feature}& & Reduced the feature space available to an adversary. & Evaluated different state of the art perturbation generation methods. & MNIST, CIFAR-10, and ImageNet & White Box & MNIST (99.43\%), CIFAR-10 (94.84\%), and ImageNet (68.36\%) & Roughly achieved 100\% for each model using different attack algorithms. & MNIST (62.7\%), CIFAR-10 (77.27\%), and ImageNet (68.11\%) \\ \cline{1-1} \cline{3-9} 
Gao et al. \cite{gao2017deepcloak}& & Proposed DeepCloak that removes unnecessary features in the model. & Perturbations are generated using FGSM. & CIFAR-10 & Not clearly articulated & 93.72\% (1\% masking) & 39.23\% (1\% masking) & 10\% increase in adversarial settings \\ \cline{1-1} \cline{3-9} 
Garg et al. \cite{garg2018spectral}& & Constructed adversarially robust features using spectral property of the dataset. & $L_2$ Perturbations & MNIST & Not clearly articulated & Not clearly articulated & Not clearly articulated & Provided empirical evidence for the effectiveness of the proposed defense. \\ \cline{1-1} \cline{3-9} 
Song et al. \cite{song2017pixeldefend}&  & Proposed PixelDefend to clean adversarial examples by moving them back to the manifold of original training data. & Used five state of the art adversarial attacks. & Fashion MNIST and CIFAR-10 & White Box & 90\% & Fashion MNIST (63\%), CIFAR-10 (32\%) for strongest defense. & Adversarial accuracy increased from 63\% to 84\% for Fashion MNIST and 32\% to 70\% for CIFAR-10. \\ \cline{1-1} \cline{3-9} 
Prakash et al. \cite{prakash2018deflecting}& & Used wavelet-based denoising method to clean natural and adversarial noise. & Generated perturbations using pixel deflection $L_2(\epsilon=0.05)$. & ImageNet & White Box & 98.9\% & Not clearly articulated & 81\% accuracy \\ \cline{1-1} \cline{3-9} 
Xie et al. \cite{xie2017mitigating}& & Proposed to use randomization at inference stage and used random resizing and random padding. & $L_\infty(\epsilon=10/255)$ & ImageNet & White Box and Black Box & 99.2\% & Not clearly articulated & 86\% accuracy \\ \cline{1-1} \cline{3-9} 
Guo et al. \cite{guo2017countering}& & Investigated different image transformation methods defending adversarial attacks. & $L_2(\epsilon=0.06)$ & ImageNet & Gray Box and Black Box & 75\% & Not clearly articulated & 70\% accuracy \\ \cline{1-1} \cline{3-9} 
Goodfellow et al. \cite{goodfellow2014explaining}& & Augmented adversarial examples into the training set. & Fast Gradient Sign (FGSM) method & MNIST & Not clearly articulated & Not clearly articulated & 99.9\% & With adversarial training, the error rate fell to 17.9\% \\ \hline 
Schott et al. \cite{schott2018towards}& \multirow{14}{11mm}{Adding Auxiliary Model} & Used generative modelling using variational autoencoder (VAE.) & Applied score-based, decision-based, transfer-based, and gradient-based attacks using $L_2(\epsilon=1.5)$. & MNIST & Not clearly articulated & 99\% & Not clearly articulated & 80\% \\ \cline{1-1} \cline{3-9} 
Wong et al. \cite{wong2018provable}& & Formulated a robust optimization problem using convex outer approximation for detection of adversarial examples. & FGSM and gradient descent based methods, $L_\infty(\epsilon=0.1)$. & MNIST & Not clearly articulated & 98.2\% accuracy & Not clearly articulated & 94.2\% accuracy \\ \cline{1-1} \cline{3-9} 
Liao et al. \cite{liao2018defense}& & Proposed a high-level representation guided denoiser (HGD) for defending adversarial attacks. & $L_\infty(\epsilon=4/255)$ & ImageNet & White Box and Black Box & 75\% & Not clearly articulated & 75\% accuracy \\ \hline
Ross et al. \cite{ross2018improving}& \multirow{24}{11mm}{Modifying Model}  & Trained the model with input gradient regularization for defending adversarial attacks. & Evaluated three famous attacks, i.e., FGSM, TGSM, and JSMA. & MNIST, Street-View House Numbers (SVHN), and notMNIST & White Box and Black Box & Not clearly articulated & Not clearly articulated & MNIST (100\%), SVHN ($\sim$90\%), and notMNIST (100\%)-approximately same for each type of attack. \\ \cline{1-1} \cline{3-9} 
Madry et al. \cite{madry2017towards}& & Trained the model with optimized parameters using robust optimization. & $L_\infty(\epsilon=8/255)$ & CIFAR-10 & White Box and Weaker Black Box & 87\% & Not clearly articulated & 46\% accuracy \\ \cline{1-1} \cline{3-9} 
Buckman et al. \cite{buckman2018thermometer}& & Proposed to use thermometer encoding for inputs. & $L\infty(\epsilon=8/255)$ & CIFAR-10 & White Box & 90\% & Not clearly articulated & 79\% accuracy \\ \cline{1-1} \cline{3-9} 
Dhillon et al. \cite{dhillon2018stochastic}& & Proposed stochastic activation pruning of the trained model for defense. & $L_\infty(\epsilon=4/255)$ & CIFAR-10 & Not clearly articulated & 83\% & Not clearly articulated & 51\% accuracy \\ \cline{1-1} \cline{3-9} 
Croce et al. \cite{croce2018provable}& & Proposed a regularization scheme for ReLU networks & Perturbations using $L_2$ and $L_\infty$ methods. & MNIST, German Traffic Signs (GTS), Fashion MNIST, and CIFAR-10. & Not clearly articulated & 98.81\% & Not clearly articulated & 96.4\% accuracy (on first 1000 test points) \\ \hline
\end{tabular}
}
\label{tab:summary_defenses}
\end{table*}

\subsection{Adversarial Defense Evaluation: Methods and Recommendations}
\label{sec:eval_method}

% \jq{Try to prepare a table or a taxonomy to accompany the text in this section.}

This section presents different potential methods for performing the evaluation of adversarial defenses along with an outline of common evaluation recommendations, as depicted in Figure \ref{fig:eval_methods}.

\subsubsection{Principles for Performing Defense Evaluations}
In a recent study \cite{carlini2019evaluating}, Carlini et al. provided recommendations for evaluating adversarial defenses and thereby provided three common reasons to evaluate the performance of adversarial defenses. These recommendations are briefly described below.

\paragraph{Defending Against the Adversary} 
Defending against adversaries attempting adversarial attacks on the system is crucial as it is a matter of security concern. In real-world applications, if the ML-based systems are deployed without considering the security threats then the adversaries willing to harm the system will continue to practice attacking the system as long as there are incentives. The nature and sovereignty of attacks vary with adversarial capabilities and knowledge, etc. In this regard, proper and well-thought threat modeling (described in detail in an earlier section) is of paramount importance.   

\paragraph{Testing Worst-Case Robustness}
In real-world settings, testing the worst-case robustness of ML models from the perspective of an adversary is crucial as real-world systems exhibit randomness that is hard to be predicted. Compared to the random testing approach, worst-case analysis can be a powerful tool to distinguish a system that fails one time in a billion trials from a system that never fails. For instance, if a powerful adversary who is attempting to harm a system to get intentional misbehavior fails to do so, then it provides strong evidence that the system will not misbehave in case of previously unforeseen randomness.    

\paragraph{Measuring Progress of ML Towards Human Level Abilities}
To advance ML techniques, it is important to understand why ML algorithms fail in a particular setting. In the literature, we see that the performance gap between ML methods and humans is considerably small on many complex tasks, e.g., natural image classification \cite{krizhevsky2012imagenet}, mastering the game of Go using reinforcement learning \cite{silver2016mastering}, and human level accuracy in the medical domain \cite{rajpurkar2017chexnet,grewal2018radnet}. However, in case of evaluating adversarial robustness, the performance gap between humans and ML systems is very large. This is so true for the cases where ML models exhibit super-human accuracy, i.e., an adversarial attack can completely evade the prediction performance of the system. This leads to the belief that there exists a fundamental difference between the decision making  process of humans and ML models. So, keeping this aspect in mind, adversarial robustness is the measure of ML progress that is orthogonal to performance.   

\subsubsection{Common Evaluation Recommendations}
In this section, we provide a brief discussion on the common evaluation recommendations and we refer interested readers to the recent article of Carlini et al. \cite{carlini2019evaluating} for a detailed and comprehensive description on evaluation recommendations and pitfalls for adversarial robustness. As authors promised to update this paper timely, therefore, we also refer interested readers to following URL\footnote{\url{https://github.com/evaluating-adversarial-robustness/adv-eval-paper}} for an updated version of this paper. To avoid unintended consequences and pitfalls of evaluation methods, the following evaluation recommendations can be adopted. 

\paragraph{Use Both Targeted and Untargeted Attacks}
Adversarial robustness should be evaluated on both targeted and untargeted attacks. In any case, it is important to explicitly state which attack were considered while evaluating. Theoretically, an untargeted attack is considered to be strictly easier than a targeted attack but practically, performing an untargeted attack can give better results than targeting any of $N-1$ classes. Many untargeted attacks mainly work by minimizing the prediction confidence of the correct label. Contrarily, targeted attacks work by maximizing the prediction confidence of some other class.

\paragraph{Perform Ablation}
Perform ablation analysis by removing a combination of defense components and verifying that the attack succeeds on a similar but undefended model. This is useful to develop a straight forward understanding of the goals of the evaluation and assess the effectiveness of combining multiple defense strategies for robustifying the model.   

\paragraph{Diverse Test Settings}
Perform the evaluation in diverse settings, i.e., test the robustness to random noise, validate broader threat models, and carefully evaluate the attack hyperparameters and select those that provide the best performance. It is also important to verify that the attack converges under selected hyperparameters. Also,  investigate whether attack results are sensitive to a specific set of hyperparameters. In addition, experiment witg at least one hard label attack and one gradient free attack.    

\paragraph{Evaluate Defense on Broader Domains}
For a defense to be truly effective, consider evaluating the proposed defense method on broader domains other than images. For instance, the majority of works on adversarial machine learning mainly investigate the imaging domain. State explicitly if the defense is only capable of defending adversarial perturbations in a specific domain (e.g., images). 

\paragraph{Ensemble Over Randomness}
It is important to create adversarial examples by ensembling over the randomness of those defenses that randomize aspects of DNN inference. The introduced randomness enforces stochasticity and standard attacks become hard to be realized. Verify that the attack remains successful when randomness is assigned a fixed value. Also, define the threat model and the availability of randomness knowledge to the adversary.  

\paragraph{Transferability Attack}
Select a similar substitute model (to the defended model) and perform transferability of the attack. Because the adversarial examples are often transferable across different models, i.e., an adversarial sample constructed for one model often appears adversarial to another model with identical architecture \cite{papernot2016transferability}. This is true regardless of the fact that the other model is trained on completely different data distribution.  

\paragraph{Upper Bound of Robustness}
To provide upper bound on robustness, apply adaptive attacks, i.e., give access to a full defense. Apply the strongest attack for a given threat model and defense being evaluated. Also, verify that adaptive attacks perform better than others and evaluate their performance in multiple settings, e.g., the combination of transfer, random-noise, and black-box attacks. For instance, Ruan et al. evaluated the robustness of DNN and presented an approximate approach to provide lower and upper bounds on robustness for $L_0$ norm with provable guarantees \cite{ruan2018global}. 

\subsection{Testing of ML Models and Autonomous Vehicles}
% \label{sec:auto_overview}

\subsubsection{Behavior Testing of Models} 
In a recent study, Sun et al. proposed four novel testing criteria for verifying structural features of DNN using MC/DC\footnote{MC/DC (Modified Condition/Decision Coverage) is a method of measuring the extent to which safety-critical software has been adequately tested.} coverage criteria \cite{sun2018testing}. They validated proposed methods by generating test cases guided by their proposed coverage criteria using both symbolic and gradient-based approach and showed that their method was able to capture undesired behaviors of DNN. Similarly, a set of multi-granularity testing criteria named DeepGauge is presented in \cite{ma2018deepgauge} that works by rendering a multi-faceted testbed. The security analysis of neural networks based system using symbolic intervals is presented in \cite{wang2018formal} which uses interval arithmetics and symbolic intervals together with other optimization methods to minimize confidence bound of over-estimation of outputs. A coverage guided fuzzing method for testing neural networks for goals (e.g., finding numerical errors, generating disagreements, and determining the undesirable behavior of models) is presented in \cite{odena2018tensorfuzz}. In \cite{guo2018dlfuzz}, the first approach utilizing differential fuzzing testing is presented for exploiting incorrect behavior of DL systems.

% \subsubsection{Mutation Testing}
% Ma et al. performed mutation testing of deep models by injecting faults to gauge the quality of test data \cite{ma2018deepmutation}. 

\subsubsection{Automated Testing of ML Empowered Autonomous Vehicles: An Overview}

To ensure a completely secure functionality of autonomous vehicles in a real-world environment, the development of automated testing tools is required. As the backbone of autonomous vehicles leverage different ML techniques for building decision systems at different levels, e.g., perception, decision making, and control, etc. In this section, we provide an overview of various studies performing test of autonomous vehicles.   

% \textcolor{red}{Because the fundamental component and backbone of autonomous vehicles leverage ML-based systems at a large level to perform different operations, e.g., perception, decision making, and control, etc.} 

% \subsubsection{Testing Deep Models for Autonomous Vehicles}
Tian et al. \cite{Tian:2018:DAT:3180155.3180220} proposed and investigated a tool named DeepTest to perform testing of DNN empowered autonomous vehicles to automatically detect erroneous behaviors of the vehicle that can potentially cause fatal accidents. Their proposed tool automatically generates test cases using changes in real-world road conditions such as weather and lighting conditions and then systematically explores different components of DNN logic that maximize the number of activated neurons. Furthermore, they tested three DNNs that won top positions in Udacity self-driving
car challenge and found various erroneous behaviors in different real-world road conditions (e.g., rain, fog, blurring, etc.) that led to fatal accidents. In \cite{zhang2018deeproad}, used a GAN-based approach to generate synthetic scenes of different driving conditions for testing autonomous cars. A metamorphic testing approach for evaluating the software part of self-driving vehicles is presented in \cite{Zhou:2019:MTD}. 

A generic framework for testing security and robustness of ML models for computer vision systems depicting realistic properties is presented in \cite{pei2017towards}. Authors evaluated the security of fifteen state of the art computer vision systems in black box setting including Nvidia's Dave self-driving system. Moreover, it has been provably demonstrated that there exists a trade-off between adversarial robustness to perturbations and the standard accuracy of the model in a fairly simple and natural setting \cite{tsipras2018robustness}. A simulation-based framework for generating adversarial test cases to evaluate the closed-loop properties of ML enabled autonomous vehicles is presented in \cite{tuncali2018simulation}. In \cite{abeysirigoonawardena2019generating}, authors generated adversarial driving scenes using Bayesian optimization to improve self-driving behavior utilizing vision-based imitation learning. An autoencoder-based approach for automatic identification of unusual events using small dashcam video and the inertial sensor is presented in \cite{li2018automatic} that can potentially be used to develop a robust autonomous driving system. Various factors and challenges impacting driveability of autonomous vehicles along with an overview of available datasets for training self-driving is presented in \cite{guo2018safe} and challenges in designing such datasets are described in \cite{uricar2019challenges}. Furthermore, Dreossi et al. suggested that while robustifying the ML systems, the effect of adversarial ML should be studied by considering the semantics and context of the whole system \cite{dreossi2018semantic}. For example, in DL empowered autonomous vehicle, not every adversarial observation might lead to harmful action(s). Moreover, one might be interested in those adversarial examples that can significantly modify the desired semantics of the whole system. 

\section{Open Research Issues}
\label{sec:research_issues}

% \jq{You should try to develop further the links with VANETs, ADS, and connected vehicles. You may see the format followed in \cite{petit2015potential} and \cite{parkinson2017cyber} and use and extend that style for this paper.}

The advancement of ML research and its state of the art performance in various complex domains, in particular, the advent of more sophisticated DL methods might be an inherent panacea to the conventional challenges of vehicular networks. However, ML/DL methods cannot be naively applied to vehicular networks that possess unique characteristics and adaption of these methods for learning such distinguishing features of vehicular networks is a challenging task \cite{ye2018machine}. In this section, we highlight a few promising areas of research that require further investigation. % In addition, the security of ML/DL models is itself an open research problem.   

%\jq{In this paper you should highlight open research issues only related to the security of ML usage in VANETs. This subsection talks about learning dynamics of VANETS, which is not per se related to the security of ML in VANETs. You may instead move this part to another place in the paper. In this section, all the issues should only be related to the main subject topic.}

%because the efficacy and robustness of various available data-driven methods such as ML-based techniques are still questionable.

\subsection{Efficient Distributed Data Storage}

In the connected vehicular ecosystem, the data is generated and stored in a distributed fashion that raises a question about the applicability of ML/DL models at a global level. As ML models are developed with the assumption that data is easily accessible and managed by a central entity, there is a need to utilize distributed learning methods for connected vehicles so that data may be scalably acquired from multiple units in the ecosystem. 

%\subsection{Decentralized Vehicular Networks}

\subsection{Interpretable ML}
Another major security vulnerability in CAVs is the lack of interpretability of ML schemes. ML techniques in general and DL techniques specifically are based on the idea of function approximation, where the approximation of the empirical function is performed using DNN architectures. Current works in ML/DL lack interpretability, which is resulting in a major hurdle in the progress of ML/DL empowered CAVs. The lack of interpretability is exploited by the adversaries to construct adversarial examples for fooling the deployed ML/DL schemes in autonomous vehicles, i.e., physical attacks on self-driving vehicles as discussed above. Development of secure, explainable, and interpretable ML techniques for security-critical applications of CAVs is another open research issue.  

\subsection{Defensive and Secure ML}
Despite many defense proposals presented in the literature for adversarial attacks, developing adversarially robust ML models remains yet another open research problem. Almost every defense has been shown to be only effective for a specific attack type and fails for stronger or unseen attacks. Moreover, most defenses address the problem of adversarial attacks for computer vision tasks but adversarial ML is being developed for many other vertical application domains. Therefore, development of efficient and effective novel defense strategies is essentially required, particularly, for safety-critical applications, e.g., communication between connected vehicles. 

\subsection{Privacy Preserving ML}
Preserving privacy in any user-centric application is of high concern. Privacy means that models should not reveal any additional information about the subjects involved in collected training data (a.k.a. differential privacy) \cite{dwork2014algorithmic}. As CAVs involve human subjects, ML model learning should be capable of preserving the privacy of drivers, passengers, and pedestrians where privacy breaches can results in extremely harmful consequences.  

\subsection{Security Centric Proxy Metrics}
Development of security-centric proxy metrics to evaluate security threats against systems is fundamentally important. Currently, there is no way to formalize different types of perturbation properties, e.g., indistinguishable and content-preserving, etc. In addition, there is no function to determine that a specific transformation is content-preserving. Similarly, the process of measuring perceptual similarity between two images is very complex and widely used perceptual metrics are shallow functions that fail to account for many subtle distinctions of human perception \cite{zhang2018unreasonable}. 
% Therefore, development of such metrics is very challenging and the best proxy approach to measure content similarity relies on DNNs \cite{zhang2018unreasonable}. 
\subsection{Fair and Accountable ML}
The literature on ML reveals that ML-based results and predictions lack fairness and accountability. The \textit{fairness} property ensures that the ML model did not nurture discrimination against specific cases, e.g., favoring cyclists over pedestrians. This bias in ML predictions is introduced by the biased training data and results in social bias and higher error rate for a particular demographic group. For example, researchers identified a risk of bias in the perception system of autonomous vehicles to recognize pedestrians with dark skin \cite{wilson2019predictive}. This is an experimental work in which authors evaluated different models developed by other academic researchers for autonomous vehicles. Despite the fact that this work does not use an actual object detection model that is being used by autonomous vehicles in the market, nor did it use the training data being used by autonomous vehicle manufactures, this study highlights a major vulnerability of ML models used in autonomous vehicles and raises serious concerns about their applicability in real-world settings where a self-driving vehicle may encounter people from a variety of demographic backgrounds.

The \textit{accountability} of ML models is associated with their interpretability property as we are interested in developing such models that can explain their predictions using the models' internal parameters. The notion of accountability is fundamentally important to understand ML model failures for adversarial examples.  

\subsection{Robustifying ML Models Against Distribution Drifts}
To restrict the integrity attacks, ML models should be made robust against distribution drifts which refer to the situation where train and test data distributions are different. This difference between the training and test distributions gives rise to adversarial examples.  These examples can also be considered as the worst case distribution drifts \cite{papernot2016towards}. It is fairly clear that the data collection process in the vehicular ecosystem is temporal and dynamic in nature so such distribution drifts are highly possible and will affect the robustness of the underlying ML systems. Moreover, such drifts can be exploited by the adversaries to create adversarial samples during inference, for example, in \cite{lowd2005good} authors investigated this distribution drift by introducing positively connotated words in spam emails to evade detection. Moreover, modification of the training distribution is also possible in a similar way and distribution drift violates the widely known presumption that we can achieve low learning error when a large training data is available. Ford at al. \cite{ford2019adversarial} have presented empirical and theoretical evidence that adversarial examples are a consequence of test error in noise caused by a distributional shift in the data. To ensure that the adversarial defense is trustworthy, it must provide defense against data distribution shifts.  As the perception system of CAVs is mainly based on data-driven modeling using historical training data, it is highly susceptible to the problem of distribution drifts. Therefore, robustifying ML models against the aforementioned distribution drifts is very important. 
One way to counter this problem is to leverage deep reinforcement learning (RL) algorithms for developing the perception system of autonomous vehicles but this is not yet practically possible, as the state and action spaces in realistic settings (road and vehicular environment) are continuous and very complex. Therefore, fine control is required for the efficacy of the system \cite{wang2018deep}. However, the work on leveraging deep RL-based methods for autonomous vehicles is building up. For instance, Sallab et al. proposed a deep RL-based framework for autonomous vehicles that enables the vehicle to handle partially observable scenarios \cite{sallab2017deep}. They investigated the effectiveness of their system using an open source 3D car racing simulator (TORCS\footnote{\url{http://torcs.sourceforge.net/}}) and demonstrated that their model was able to learn complex road curvatures and simple inter-vehicle interactions. On the counter side, deep RL-based systems have been shown vulnerable to policy induction attacks \cite{behzadan2017vulnerability}.  

\section{Conclusions}
\label{sec:con}

%\textcolor{red}{Original equipment manufacturers are expecting to launch autonomous vehicles for commercial use by 2020. There is no doubt that connected and autonomous vehicles (CAVs) represent promising technologies that have much potential. However, CAVs introduce various technical and social challenges including the prominent challenge relating to how to efficiently, accurately, and robustly apply machine learning (ML)/deep learning (DL) techniques which are extensively used for diverse vehicular tasks that range from perceiving the environment to predicting actions and recognizing objects. Recently, these techniques have been found to be strictly vulnerable to carefully crafted adversarial examples. This raises many concerns about the security of ML/DL models and by extension of the security of ML/DL-enabled CAVs. In this work, we present for the first time a comprehensive analysis of the challenges posed by adversarial ML attacks on CAVs. Towards this end, we present the ML pipeline of CAVs and then present various adversarial attacks that can be realized in practical settings. Then, we present an outline of potential solutions to defend against adversarial attacks. Further, we highlight that without necessary robustness against the security threats, ML models in CAVs can cause more harm than good. Finally, open research challenges and future directions are discussed to provide readers with the opportunity to develop robust and efficient solutions for the application of ML models in CAVs.}

The recent discoveries that machine learning (ML) techniques are vulnerable to adversarial perturbations have raised questions on the security of connected and autonomous vehicles (CAVs), which utilize ML techniques for various tasks ranging from environmental perception to objection recognition and movement prediction. The safety-critical nature of CAVs clearly demands that the technology it uses should be robust to all kinds of potential security threats---be they accidental, intentional, or adversarial. In this work, we present for the first time a comprehensive analysis of the challenges posed by adversarial ML attacks for CAVs aggregating insights from both the ML and CAV literature. Our major contributions include: a broad description of the ML pipeline used in CAVs; description of the various adversarial attacks that can be launched on the various components of the CAV ML pipeline; a detailed taxonomy of the adversarial ML threat for CAVs; a comprehensive survey of adversarial ML attacks and defenses proposed in literature. Finally, open research challenges and future directions are discussed to provide readers with the opportunity to develop robust and efficient solutions for the application of ML models in CAVs.

%Towards this end, we present the ML pipeline of CAVs and then present various adversarial attacks that can be realized in practical settings. Then, we present an outline of potential solutions to defend against adversarial attacks. Further, we highlight that without necessary robustness against the security threats, ML models in CAVs can cause more harm than good. 

%a ML pipeline of self-driving VANETs and describes potential security issues related to ML/DL techniques used in VANETs.  
%\jq{Conclude the paper here. Make sure that you highlight the contributions of this paper and how it extends the state of the art.}

\bibliographystyle{unsrt}
%\bibliographystyle{plain}
% \bibliography{vehicular}

% \end{document}

\bibliographystyle{plain}
%\bibliography{references}

% that's all folks
\end{document}